\documentclass[pdflatex,sn-basic]{sn-jnl}

\usepackage{natbib}

\usepackage{graphicx}         
\usepackage{afterpage}         
\usepackage{numprint}         
\npstyleenglish

\usepackage{bm}
\usepackage{bbm} 
\usepackage{amsthm,amssymb,amsmath}

\usepackage{multirow}
\usepackage{array}

\usepackage{makecell}
\newcolumntype{P}[1]{>{\centering\arraybackslash}p{#1}}
\usepackage{algorithm}

\usepackage{algpseudocode}
\usepackage{arydshln}
\usepackage{url}

\jyear{2023}%

\theoremstyle{thmstyleone}%
\theoremstyle{thmstyletwo}%

\theoremstyle{thmstylethree}%

\begin{document}

\title[Monte Carlo sampling of valid and realistic counterfactual explanations]{MCCE: Monte Carlo sampling of valid and realistic counterfactual explanations for tabular data}


\author{\fnm{Annabelle} \sur{Redelmeier}}\email{aredelmeier@gmail.com}

\author*{\fnm{Martin} \sur{Jullum}}\email{martin.jullum@nr.no}

\author{\fnm{Kjersti} \sur{Aas}}\email{kjersti.aas@nr.no}

\author{\fnm{Anders} \sur{Løland}}\email{anders.loland@nr.no}

\affil{\orgname{Norwegian Computing Center}, \orgaddress{\street{P.O. Box 114}, \city{Oslo}, \postcode{NO-0314}, \country{Norway}}}


\abstract{We introduce MCCE: \textbf{\underline{M}}onte \textbf{\underline{C}}arlo sampling of valid and realistic \textbf{\underline{C}}ounterfactual \textbf{\underline{E}}xplanations for tabular data, a novel counterfactual explanation method that generates on-manifold, actionable and valid counterfactuals by modeling the joint distribution of the mutable features given the immutable features and the decision. 
  Unlike other on-manifold methods that tend to rely on variational autoencoders and have strict prediction model and data requirements, MCCE handles any type of prediction model and categorical features with more than two levels. 
  MCCE first models the joint distribution of the features and the decision with an autoregressive generative model where the conditionals are estimated using decision trees. Then, it samples a large set of observations from this model,
  and finally, it removes the samples that do not obey certain criteria. 
We compare MCCE with a range of state-of-the-art on-manifold counterfactual methods using four well-known data sets and show that MCCE outperforms these methods on all common performance metrics and speed.
In particular, including the decision in the modeling process improves the efficiency of the method substantially.}


\maketitle

\section{Introduction}\label{sec:Intro}

It is exceedingly apparent that complex, black-box AI, machine learning, and statistical models deployed in industry need sound and efficient explanations. In this paper, we discuss counterfactual explanations, a type of local prediction explanation method. 
Counterfactual explanations (CEs)\footnote{We use `counterfactual explanation' or `CE' to refer to the literature or explanation type and `counterfactual' or `example' to refer to the instance produced.} explain predictions, $f(\bm{x})$, from a fitted prediction model $f(\cdot)$ or some other deterministic ML/AI system. It does so by providing examples that yield a different \textit{decision} than the feature vector, $\bm{x}$, one is trying to explain\footnote{\label{decision_footnote}A \textit{decision} is derived from a \textit{prediction}, $f(\bm{x})$, using a pre-defined cutoff value or interval $c$, characterizing the desired decision. For example, if $f(\bm{x}) = 0.39$ and $c = (0.5, 1]$, then since $f(\bm{x}) \notin c$, we give instance $\bm{x}$ a decision of 0 and say $\bm{x}$ has received an undesirable decision.}.
These examples, $\tilde{\bm{x}}$, are called \textit{counterfactuals}, because they give an idea of what features could be changed to obtain a different decision. 
Generating a feature vector with a different decision is sometimes referred to as generating \textit{recourse} \citep{karimi2022survey,guidotti2022counterfactual}. 

As a simple example, imagine that a bank utilizes a black-box machine learning model to decide whether or not an individual should receive a loan. The model takes the four features \texttt{age}, \texttt{sex}, \texttt{salary}, and \texttt{defaulted\_last\_year} as inputs,  and outputs the probability of defaulting on the loan $\in [0,1]$. If the probability is in the desired decision interval $c = [0, 0.1)$, the bank customer is granted the loan.
Now suppose that a 36-year-old female customer with a salary of \$68,000 who defaulted on a loan last year, is denied a loan. 
To \textit{explain} to the customer why her loan was denied, a set of counterfactuals could be provided. 
One counterfactual may show that if she had a \texttt{salary} of \$80,000 rather than \$68,000  she would have received the loan. 
Another may show that only increasing her salary to \$72,000, but without having defaulted on her loan last year would also make her eligible for the loan. 

For counterfactuals to be useful, they should be of low \textit{cost}, i.e.,  limit the \textit{number} and \textit{magnitude} of feature changes. 
Low-cost counterfactuals are easier for individuals to implement and/or case handlers to understand.
Counterfactuals should also be \textit{actionable}, i.e., not change features that are fixed by the individual. Otherwise, they will be impossible to implement in practice. Further, the counterfactuals should lie on the data manifold, since this ensures a realistic and plausible combination of features.
Finally, counterfactuals should be \textit{valid}, i.e., obtain the desired decision. Without this, the counterfactual would not be a \textit{counter}factual at all.  

\subsection{Related work}
Recently there has been an explosion of papers proposing counterfactual explainers. Surveys are given by e.g.,\ \cite{Verma21}, \cite{Stepin21} and \cite{guidotti2022counterfactual}. In the most recent survey \citep{guidotti2022counterfactual},
the different approaches are categorized according to (i) the strategy used to generate the counterfactuals, (ii) whether the approach is model-agnostic, (iii) whether the approach can be used for tabular data, text or images, (iv) whether the approach is able to handle categorical data,
(v) whether the approach guarantees validity, and finally,  (vi) whether the approach is able to ensure actionability. The survey shows that most counterfactual explanation approaches are tailor-made for either tabular data, text, or images.
The strategy used to generate the counterfactuals is categorized into four groups: (i) optimization-based, (ii) heuristic search-based, (iii) decision-tree-based, and (iv) instance-based. The majority of the approaches belong to
the first two categories. In what follows we describe and briefly review the four groups of approaches.  

The optimization-based approaches usually first define a loss function of the form 
\begin{equation}\label{eq:wachter}
    \text{dist}_1(f(\tilde{\bm{x}}), y') + \lambda \cdot \text{dist}_2(\bm{x}, \tilde{\bm{x}}),
\end{equation}
where $f(\cdot)$ is a prediction model, $\bm{x}$ is the original feature vector, $\bm{\tilde{x}}$ is the counterfactual, $\lambda$ is some tuning parameter, $y' = \mathbbm{1}(f(\bm{\tilde{x}})\in c)$ is an indicator for the desired decision (see footnote \ref{decision_footnote} on page \pageref{decision_footnote}),
and $\text{dist}_1(\cdot,\cdot)$ and $\text{dist}_2(\cdot,\cdot)$ are two appropriate distance functions (e.g., weighted $L_0$, $L_1$, $L_2$ norms or the median absolute deviation). The counterfactual $\bm{\tilde{x}}$ is then found by e.g.,\
stochastic gradient descent \citep{wachter2017counterfactual}, integer programming \citep{ustun2019actionable}, random walks \citep{laugel2017inverse}, genetic algorithms \citep{dandl2020multi, rasouli2022care}, or through a sequence of satisfiability (SAT) problems \citep{karimi2020model}. 
The optimization-based approaches often restrict $f(\cdot)$ to be differentiable, meaning that they do not work for tree-based classifiers like XGBoost 
\citep{chen2016xgboost} or random forest \citep{breiman1984classification}. Further, these methods usually cannot guarantee that the counterfactual will be actionable, and most of them do not handle categorical features with more than two levels.
Finally, most of the optimization-based methods produce counterfactuals that do not lie on the data manifold \citep{pawelczyk2021carla}. There are, however, some exceptions. The methods \texttt{CEM-VAE} \citep{dhurandhar2018explanations},
\texttt{REViSE} \citep{joshi2019towards}, \texttt{CLUE} \citep{antoran2020getting}, \texttt{C-CHVAE} \citep{pawelczyk2020learning}, and \texttt{CRUDS} \citep{downs2020cruds} all use variational autoencoders (VAE) to find counterfactuals
that are proximal and connected to input data.

The approaches that use heuristic search strategies, of which two examples are \texttt{VICE} \citep{Gomez20} and \texttt{MOC} \citep{dandl2020multi}, are quite similar to the optimization-based ones, but instead of using optimization
to minimize a loss function, heuristic search strategies are used. Such strategies are usually much
faster than optimization algorithms. The downside is that the solutions are not necessarily optimal. Moreover, the heuristic search-based methods have many of the same weaknesses as the optimization-based approaches listed above.

The decision-tree based methods approximate the behavior of the prediction model with a decision tree and then exploit the logic revealed by the tree for building counterfactual explanations. One example is \texttt{FT} \citep{Tolomei17}. A weakness of this approach is that the logic learned by the surrogate decision tree need not be the same as that of the original prediction model. 

The last category, instance-based approaches, retrieves counterfactuals by selecting the most similar examples from a (simulated) data set. The approach proposed in this paper belongs to this category. According to \cite{guidotti2022counterfactual}
the instance-based methods are the best performers with respect to all the properties benchmarked in their survey. Four instance-based methods are discussed in \cite{guidotti2022counterfactual}. Three of these methods, \texttt{CBCE} \citep{Keane20},
\texttt{NICE} \citep{brughmans2023nice}, and \texttt{FACE} \citep{poyiadzi2020face}, do not treat immutable features in an appropriate way. 
The fourth approach, \texttt{NNCE} \citep{Wexler20}, should not be used for privacy reasons, because it selects counterfactuals as actual rows in the training data set. 

As shown above, every existing method presents one or more limitations. Given that the instance-based approaches seem to be the most productive, we believe there is a demand for a method of this kind that successfully addresses and resolves the identified shortcomings.

\subsection{Our contribution}

In this paper, we introduce MCCE (\textbf{\underline{M}}onte \textbf{\underline{C}}arlo sampling of valid and realistic \textbf{\underline{C}}ounterfactual \textbf{\underline{E}}xplanations for tabular data), a novel instance-based method that
handles any type of prediction model and feature cardinality.
MCCE consists of three main steps. In step 1, MCCE models the data distribution and the decision using an autoregressive generative model, e.g., MADE \citep{germain2015made}, with conditionals estimated using decision trees.
In step 2, MCCE generates a large set of samples that respect the learned data manifold using the model from step 1. Finally, in step 3, MCCE obtains counterfactual explanations by extracting the low-cost and valid samples.

MCCE is different from previously proposed CE methods in that it:
\begin{enumerate}
\item  Uses a simultaneous model for the features \textit{and} the decision to ensure on-manifold and valid counterfactuals. It models the decision through an additional fixed binary feature $y' = \mathbbm{1}(f(\bm{x})\in c)$,
for a desired decision interval $c$ (see footnote \ref{decision_footnote} on page \pageref{decision_footnote}). When generating samples in step 2, MCCE conditions on $y' = 1$, which guides the samples to the correct decision. This increases the efficiency of the method substantially.  To the best of our knowledge, this trick has not been exploited by CE methods before.
\item  Generates examples that are guaranteed to obey the immutable features (e.g., age, sex), i.e., the samples are actionable. This guarantee stems from the way the model in step 1 is specified. 
\item  Uses decision trees to estimate the data distribution. This means that MCCE does not require the underlying prediction model to be a gradient-based classifier like CE methods that use VAEs. 
\item Handles categorical features with an arbitrary number of levels. 
\item Breaks up the task into independent steps that can easily be altered to target specific needs. 
\end{enumerate}

The rest of the paper is organized as follows. 
In Section \ref{sec:MCCE}, we describe our method, MCCE, in detail.
In Section \ref{sec:real_data}, we compare the counterfactual explanations from MCCE with those of various competing methods and show how MCCE surpasses all other methods when it comes to speed and performance metrics. We also investigate how sensitive MCCE is to its main tuning parameter, study the characteristics of the data generated in the second step of the method, and examine the scalability of the different steps of the MCCE procedure. Section \ref{privacy} examines the privacy implications of MCCE before we conclude in Section \ref{sec:conclusion}. 

\section{Generating counterfactual explanations with MCCE}\label{sec:MCCE}

We now present MCCE, our algorithm for generating counterfactuals. Suppose that we have a tabular data set $\bm{X} \in \mathcal{X}$ with dimension $n_{\text{train}} \times p$ and binary response $\bm{Y} \in \mathcal{Y}$. 
Suppose further that we have a prediction model $f(\cdot)$, a set of predictions, $f(\bm{x}_i)$, and a set of observations $H$ that have obtained an undesirable decision, i.e., $H := \{i: f(\bm{x}_i) \notin c\}$ for decision interval $c$, that we wish to explain. 
Note that given $\bm{x} \in \mathcal{X}$, $f(\bm{x})$ is deterministic, i.e., always produces the same prediction.
We divide the feature space, $\mathcal{X}$, into mutable features $\mathcal{X}^m$ and immutable features $\mathcal{X}^f$ such that $\mathcal{X} = \mathcal{X}^m \times \mathcal{X}^f$. 
Then, for each observation $h \in H$ with factual $\bm{x}_h$, our aim is to create a set of $J$ counterfactuals $\bm{E}_h = \{{\bm{e}}_{h,1},\ldots, {\bm{e}}_{h,J}\}$, such that:
\begin{itemize}
    \item \textbf{Criterion 1}: $\bm{e}_{h,j}$ is \textit{on-manifold}, i.e., $p(\bm{X}^m = {\bm{e}}^m_{h,j} \mid \bm{X}^f = {\bm{e}}^f_{h,j})>\epsilon$, for some $\epsilon > 0$;
    \item \textbf{Criterion 2}: $\bm{e}_{h,j}$ is \textit{actionable}, i.e., does not violate any of the immutable features; 
    \item \textbf{Criterion 3}: $\bm{e}_{h,j}$ is \textit{valid}, i.e., $f(\bm{e}_{h,j}) \in c$, for the desired decision interval $c$;
    \item \textbf{Criterion 4}: $\bm{e}_{h,j}$ is \textit{low cost}, i.e., close to the factual, $\bm{x}_h$.
\end{itemize}

Given a training data set $\bm{X}$ and these criteria, we propose to generate a set of counterfactuals with three steps:
\begin{enumerate}
    \item Model the distribution of the mutable features given the immutable features and the decision, $y'$. In our implementation, we use an autoregressive generative model based on decision trees.
    \item For each observation $h \in H$, generate a large data set of dimension $K \times (p+1)$ by sampling from this fitted distribution. Denote this data set ${\bm{D}}_h$ for each $h \in H$.
    \item Create ${\bm{E}}_h$ for each $h \in H$ by removing the rows in ${\bm{D}}_h$ that do not obey criteria 3 and 4. Return ${\bm{E}}_h \subseteq {\bm{D}}_h$ as the set of counterfactual explanations for this instance.
\end{enumerate}

\begin{figure*}
    \centering
    \includegraphics[width=1\linewidth]{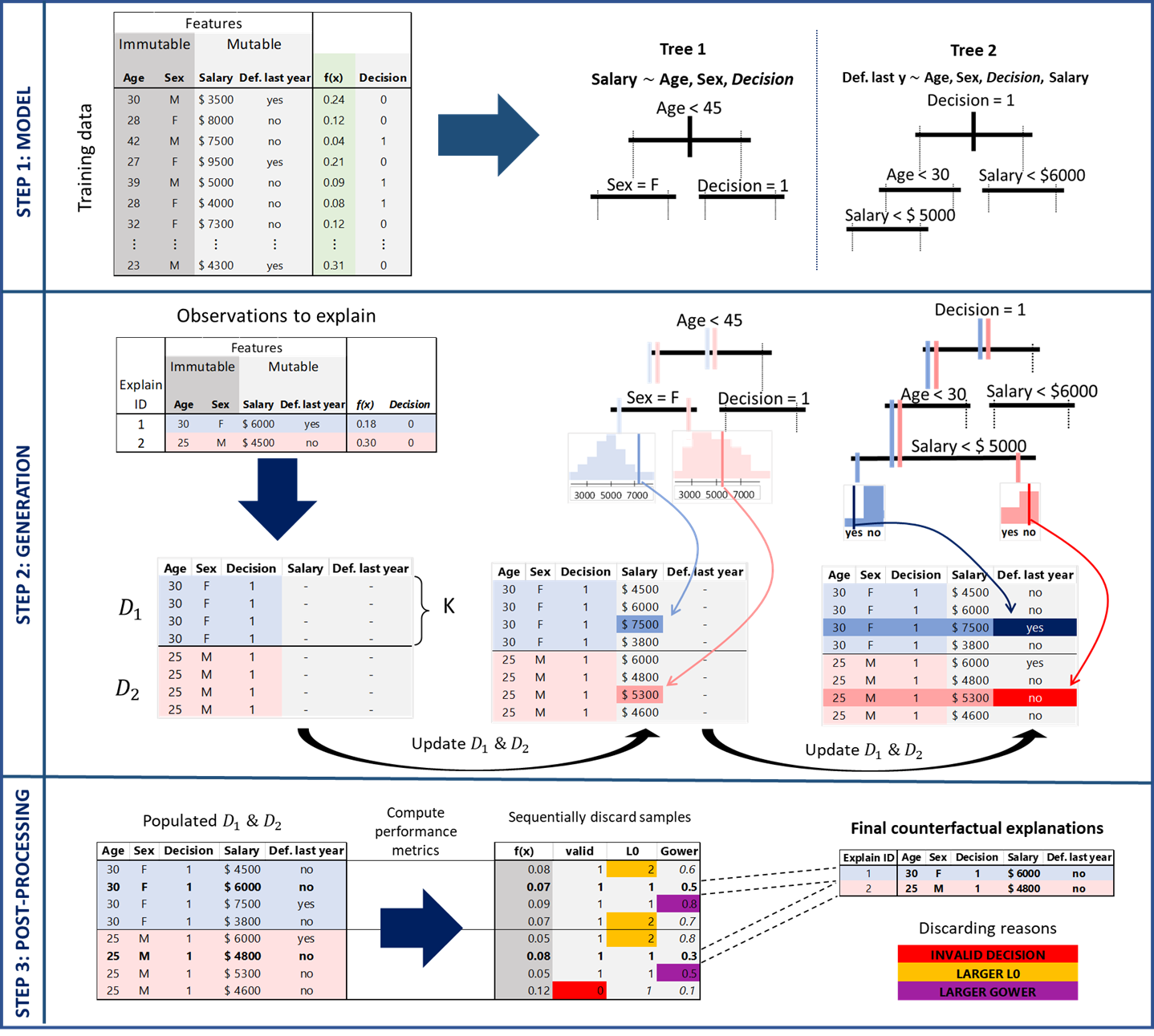} 
    \caption{MCCE's three steps illustrated on the introductory example with two immutable and two mutable features. 
    Step 1: Fit a decision tree for each mutable feature iteratively on the previously fit features and the decision. 
    Step 2: For each observation/prediction to explain, trace the trees down based on the immutable feature values, decision = 1, and previous sampled values, 
    and then randomly sample a training observation in the leaf nodes and update $\bm{D}_h$.
    Step 3: Among the sampled rows, find the instance closest to the original vector over the decision boundary.}
    \label{fig:mcce_steps}
\end{figure*}

Figure \ref{fig:mcce_steps} shows an outline of the three steps, and Sections \ref{subsec:model}-\ref{subsec:postprocessing} describe them in more detail.

\subsection{Model}\label{subsec:model}

\textbf{Step 1} (\textit{model}) 
To model the distribution of the $q$ mutable features, $\bm{X}^m = \{ {X}^m_1,\ldots, {X}^m_q \}$, given the $u$ immutable features, $\bm{X}^f = \{ {X}^f_1, \ldots, {X}^f_u \}$, and decision, $Y'$, we use an autoregressive generative model to decompose $\bm{X}^m \mid \bm{X}^f, Y'$ into products of conditional probability distributions as follows
\begin{equation}\label{eq:chain_rule}
p(\bm{X}^m \mid \bm{X}^f, Y')  =  
p(X_1^m \mid \bm{X}^f, Y') \prod_{i=2}^q p(X_i^m \mid \bm{X}^f, Y', X_1^m, \dots, X_{i-1}^m).
\end{equation}
Then, we estimate each of the univariate conditional distributions in \eqref{eq:chain_rule} separately.
The main reason for rewriting $p(\bm{X}^m \mid \bm{X}^f, Y')$ in this way is that it is typically much easier to correctly model distributions with only a single dependent variable, rather than many. 

The question is then how to model each of these conditional distributions. 
Options include the use of basic parametric probability distributions, copulas \citep{sklar1959fonctions}, non-parametric methods, and deep learning approaches such as MADE \citep{germain2015made}.
We choose to model the $q-1$ conditional distributions in \eqref{eq:chain_rule} using $q-1$ decision trees.
For example, to model $X^m_2 \mid \bm{X}^f, Y', X^m_1$, we fit a decision tree to $X^m_2 \sim (\bm{X}^f, Y', X^m_1)$ based on the data.
We use decision trees because
they are fast and robust to extrapolation, scale well, handle mixed (continuous and categorical) data automatically, handle arbitrary orders of interactions between the features, and do not require extra preprocessing or scaling of the data. 

Inspired by \cite{drechsler2011empirical} and \cite{reiter2005using}, we choose to use CART (categorical and regression trees) \citep{breiman1984classification} to model these conditional distributions. 
CART is a tree algorithm that builds trees by recursively making binary splits on the feature space until a given stopping criterion is fulfilled. 
Depending on the class of the response, either the Gini impurity index or the squared-error loss can be used to measure the quality of the split.  
The tree can be controlled by setting other parameters like the maximum depth or number of features to check when splitting a node.

We choose CART over other non-parametric models since CART is consistent \citep{scornet2015consistency, chi2022asymptotic} and allows for high dimensionality.
In addition, CART has previously been used for synthetic data generation with good results. 
For example, \cite{reiter2005using} uses CART to impute sensitive data and then compares the generated observations with the true observations. 
\citeauthor{reiter2005using} show that CART generates data that are close to the corresponding population statistics and covers the corresponding data intervals. 
Further, \cite{drechsler2011empirical} compare different non-parametric approaches to generating data. They show that CART is one of two methods that gives the best results when it comes to recreating population means and variances.

\subsection{Generation}\label{subsec:generating}

\textbf{Step 2} (\textit{generation})
Suppose that in step 1 we fit $q-1$ decision trees: $T_1, \dots, T_{q-1}$ where $T_i$ is fit to $X^m_i \sim (\bm{X}^f, Y', X_1^m, \dots, X_{i-1}^m)$. Step 2 consists of generating $K \times p$ 
dimensional data sets, $\bm{D}_h$, $h \in H$, with samples from the fitted trees.
The full generation procedure is outlined in Algorithm \ref{alg:MCCE_gen2}, but in brief, for each observation, $h$, we loop over each mutable feature, $j$, and sample a set of feature values that we append to $\bm{D}_h$. To obtain the samples for feature $j$, we find the end node of the fitted tree $T_j$ based on the current row values of $\bm{D}_h$, and sample from this node.  

\begin{algorithm}[ht]
    \caption{Outline of step 2 in MCCE}
    \label{alg:MCCE_gen2}
    \begin{algorithmic}[1]
    \Require{ \mbox{ } \newline $u > 0$ number of immutable features
    \newline $q > 0$ number of mutable features 
    \newline ($p = u+q$ is the total number of features)
    \newline $H > 0$ number of observations to explain
    \newline $K > 0$ number of samples from each tree 
    \newline $T_1, \dots, T_{q-1}$ fitted decision trees
    \newline}
        \For{$h \in H$}
            \State Let $\bm{D}_h$ be a $K \times u$ matrix where each row is a copy of the vector of $u$ immutable features $\bm{x}^f_h$
            \State Append a $K \times 1$ column vector to $\bm{D}_h$ with $y' = 1$ repeated $K$ times
            \For{$1 \le j \le q-1$}
                \State Let $\bm{d}$ be an empty vector of length $K$
                \For{$1 \le i \le K$}
                    \State Find the end node of tree $T_j$ based on vector $\bm{D}_h[i,]$.
                    \State Let $\bm{d}[i]$ be a single (Monte Carlo) sample from the training observations belonging to this end node.
                \EndFor
            \State $\bm{D}_h \gets [\bm{D}_h\, \mid \, \bm{d}]$;  \Comment{Append vector $\bm{d}$ as a new column of $\bm{D}_h$}
            \EndFor
        \State Remove the $y' = 1$ column vector from $\bm{D}_h$
        \State \textbf{return} $\bm{D}_h$
        \EndFor
         \newline
    \Ensure{\mbox{ } 
    \newline data sets $\bm{D}_h$ of dimension $K \times p$}
    \end{algorithmic}
\end{algorithm}

While the feature values in $\bm{D}_h$ only take the values of the features in the training data, our method \textit{combines} the feature values in new ways such that the rows of $\bm{D}_h$ are rarely observed in the training data. See Section \ref{privacy} for a discussion on privacy concerns. Note that in practice, Algorithm \ref{alg:MCCE_gen2} is made more efficient by only following the trace down the branches of $T_j$ for each \textit{unique} row of $\bm{D}_h$ and then returning multiple samples from that end node (one for each duplicate row). This saves time in the first stages of the algorithm. Observations $h \in H$ with the same immutable feature values can also share the same data set, $\bm{D}_h$.

\subsection{Post-processing}\label{subsec:postprocessing}

\textbf{Step 3} (\textit{post-processing}) 
The last step of MCCE removes the rows of $\bm{D}_h$ that do not satisfy the criteria listed at the start of Section \ref{sec:MCCE}. The rows that remain become the set of counterfactuals for observation $h \in H$. Criteria 1 and 2 are already fulfilled since the samples come from an approximation to the data distribution conditioned on the immutable features. 
In addition, most samples also fulfill criterion 3 since we condition on the decision when building the trees. Nevertheless, we remove the few rows of $\bm{D}_h$ where $f(\bm{D}_h[i,]) \notin c$. 

To fulfill criterion 4, 
we calculate \textbf{sparsity} and the \textbf{Gower distance} between each $\bm{D}_h[i,]$ and factual $\bm{x}_h$.
For factual $\bm{x} = \{x_1, \dots, x_p\}$ and row  $\bm{d} = \{d_{1}, \dots, d_{p} \}$, sparsity is equal to the number of features changed between $\bm{x}$ and $\bm{d}$ and
\begin{equation}\label{eq:distance}
    \text{Gower distance} = \frac{1}{p} \sum_{j=1}^p \delta_G{(d_j, x_j)} \in [0,1], 
\end{equation}
where
\begin{equation}\label{eq:gower}
    \delta_G(d_j, x_j) =
    \begin{cases} 
      \frac{1}{R_j} \mid d_j - x_j \mid & \text{if } x_j \text{ is numerical,} \\
      \mathbbm{1}_{d_j \ne x_j} & \text{if } x_j \text{ is categorical,}
   \end{cases} 
\end{equation} 
and $R_j$ normalizes the $j$th feature so that it lies between 0 and 1. 
Although we could have used a distance function other than the Gower distance, it is our impression that the Gower distance is the most commonly used distance function in the counterfactual explanation literature,
see e.g.,\ \cite{wachter2017counterfactual}, \cite{mothilal2020explaining} and \cite{guidotti2022counterfactual}.

To generate one counterfactual for each $h \in H$, we find the minimum sparsity across all instances of $\bm{D}_h$ and remove the instances with a sparsity larger than this value. 
Then we find the instance in the remaining set with the smallest Gower distance. This becomes the counterfactual for observation $h \in H$. 

In this paper, we concentrate on generating a single counterfactual per instance to be explained. 
However, if we want to use our method to return several (say $l$) counterfactuals for each $h \in H$, we can set upper bounds for sparsity and the Gower distance and present the best $l$ instances in $\bm{D}_h$ with a sparsity and Gower distance less than these bounds. See Section \ref{sec:conclusion} for further discussion on this subject.



\section{Experiments}\label{sec:real_data}

In this section, we perform four experiments. In the first experiment, we compare MCCE to a set of previously proposed methods for generating counterfactual explanations using four well-known real data sets.
The second investigates how MCCE's performance varies with the tuning parameter $K$, while the third studies the characteristics of the data generated in the second step of the algorithm. Finally, the fourth
experiment considers the scalability of the different steps of the MCCE procedure.
See also Appendices \ref{non-binarised} and \ref{non-gradient} for additional results for non-binarized data and a non-gradient based prediction model, and for a study of the importance of conditioning on the decision.

All computations in this section are run on a desktop computer with a 16-core AMD Ryzen Threadripper 1950X, 3.4 GHz CPU, and an NVIDIA GeForce GTX 1080 Ti GPU, which runs Ubuntu 20.04 with Python 3.7. MCCE runs on CPU only, while most of the competing methods can benefit from being run on the GPU. Therefore, below, the competing methods are run using a GPU (GeForce GTX 1080 Ti) whereas MCCE is run on a CPU. 

\subsection{Experiment 1: How does MCCE compare with competing methods?}\label{subsec:experiment1}

In this section, we compare MCCE to a set of previously proposed methods for counterfactual explanations using four well-known data sets.

\textbf{Data sets}
\textit{The Adult data set} is from the 1994 Census database consisting of four continuous and eight categorical features and \numprint{49000} observations. The goal is to classify individuals with an income of more than \$\numprint{50000} USD per year or not. We explain the predictions of individuals who are predicted to not reach this income threshold. 
The features \texttt{age} and \texttt{sex} are set as immutable.

\textit{The Give Me Some Credit (GMC) data set} is from a 2011 Kaggle Competition in credit scoring consisting of \numprint{150000} observations and 10 continuous features. The goal is to classify individuals as experiencing financial distress or not. Individuals with financial distress are given a response of 0 and those without are given a response of 1. We explain the predictions of individuals who are predicted to experience financial distress. We drop the rows with missing data and set \texttt{age} as immutable. 

\textit{The FICO data set} is from the 2018 FICO Explainable Machine Learning Challenge. The data set is of Home Equity Line of Credit (HELOC) applications made by homeowners. The response is a binary feature called RiskPerformance
that takes the value 1 (`Bad') if the customer is more than 90 days late on his or her payment and 0 (`Good') otherwise. There are 23 features (21 continuous and two categorical) and \numprint{10459} observations. We explain those predicted to be `Bad' and set the feature \texttt{ExternalRiskEstimate} as immutable.

\textit{The German Credit Data set} is a publically available data set downloaded from the UCI Machine Learning Repository.  It consists of 20 features (seven are continuous and 13 are categorical) and has 1000 observations. Each entry represents
a person who takes credit at a bank. Each person is classified as having good or bad credit risk according to the set of attributes. 
We explain the predictions of individuals who are predicted to have bad credit risk.
The features \texttt{Purpose}, \texttt{Age}, and \texttt{Sex} are set as immutable.

\textbf{Competing methods} We compare MCCE with the instance-based method FACE \citep{poyiadzi2020face}, and the optimization-based methods
\texttt{C-CHVAE} \citep{pawelczyk2020learning},
\texttt{CEM-VAE} \citep{dhurandhar2018explanations},
\texttt{CLUE} \citep{antoran2020getting},
\texttt{CRUDS} \citep{downs2020cruds}, and
\texttt{REViSE} \citep{joshi2019towards}.
The specific optimization-based methods were chosen because they produce counterfactuals that lie on the data manifold. In our opinion, this is an important requirement for obtaining realistic counterfactuals.
The Python package \texttt{CARLA} \citep{pawelczyk2021carla} is used to generate counterfactuals from the selected methods. For all four data sets, we use the default parameter values in \texttt{CARLA} for all the competing methods,
except for \texttt{CLUE} where we use other values for the FICO and German Credit Data sets such that the method is able to generate counterfactuals for a larger proportion of the individuals.
The parameter values are shown in Appendix \ref{parValues}. The counterfactuals were generated with code from commit 192 in \texttt{CARLA}.


\textbf{Prediction model}
For all four data sets, we use a multi-layer perceptron (ANN) with three hidden layers as the prediction model. The number of hidden nodes, epochs, and batch sizes for each data set are shown in Table \ref{tab:ann} together
with the out-of-sample AUC values. The ANN is chosen instead of e.g.,\ XGBoost or Random Forest, since with the exception of \texttt{C-CHVAE}, all the competing methods require gradient-based classifiers. 
Before fitting the models, the categorical features are binarized by partitioning them into the most frequent class and its counterpart, since most of the competing methods do not handle categorical
features with more than two levels. MCCE was also tested with non-binarized features and a random forest model in addition to the experiments described in this section. See Appendices \ref{non-binarised} and
\ref{non-gradient} for the results. All continuous variables are scaled to have mean equal to 0 and standard deviation equal to 1 before fitting the prediction models.
Finally, for the sake of simplicity, the desired decision interval $c$ is chosen to be $[0.5,1]$ for all data sets.

\begin{table}[ht!]
    \caption{Predictive model settings.}
    \centering
    \begin{tabular}{l|r|r|r|r}
         \hline
         & Adult & GMC & FICO & German Credit\\
         \hline
         Hidden nodes  & (18, 9, 3) & (18, 9, 3) & (81, 27, 3) & (81, 16, 3)\\
         Batch-size    &   1024 &  2048 &      8 &    16\\
         Epochs        &    20  &    20 &     20 &    20\\
         Learning rate & 0.002  & 0.002 & 0.0005 & 0.002\\
         \hline
         AUC for test set  & 0.90 & 0.82 & 0.76 & 0.80 \\
         \hline
    \end{tabular}
    \label{tab:ann}
\end{table}

\textbf{Our Method} For MCCE, we fit the conditional distributions in \eqref{eq:chain_rule} with CART. We use the Gini impurity index to measure the quality of the split if the `response' feature is categorical and the mean squared error if it is continuous. To generate deep trees, we set the minimum number of samples required to split to 2 and the minimum number of samples in each leaf node to 5. We do not set a max depth of the tree.  These selections closely resemble the default values for CART models in the widely-used synthetic data generating \verb|R|-package \verb|synthpop| \citep{nowok2016synthpop}.
Unless otherwise noted, we use $K = 1000$ in step 2 of the algorithm. 

\textbf{Performance metrics} There is no standard agreement on how to perform an objective evaluation of methods generating counterfactual explanations \citep{guidotti2022counterfactual}. However, like \cite{pawelczyk2021carla} and
\cite{guidotti2022counterfactual} we think it is most natural to evaluate the performance of counterfactual explainers with respect to the four criteria listed in Section \ref{sec:MCCE}. The counterfactuals should (i) lie on the data manifold,
(ii) be valid, (iii) be actionable, and (iv) be of low cost.

\textit{Data manifold closeness} is the property that guarantees that the counterfactual is as similar to the training data as possible \citep{wachter2017counterfactual}.
We follow \cite{guidotti2022counterfactual} and say a counterfactual 
lies on the data manifold if it is `similar' to the training data set and obeys the observed correlations among the features. 
A counterfactual is \textit{valid} if the counterfactual produces the correct decision (i.e., \textit{success} = 1).
A counterfactual is \textit{actionabile} if it has as few violations as possible. A violation is when an immutable feature is changed.
Finally, a counterfactual is of \textit{low-cost} if it minimizes sparsity (denoted $L_0$) and the Gower distance (denoted $L_1$), see Section \ref{subsec:postprocessing} for definitions.

\textbf{Results} 
For each data set, except the German Credit Data set, we generate one counterfactual for $n_\text{test} = 1000$ observations with an undesirable decision using MCCE and the competing methods. The German Credit Data set only has 1000 observations in total; therefore, we generate counterfactuals for 200 observations.  We denote the number of counterfactuals generated by each method as $N_{\text{CE}}$.
If $N_{\text{CE}} < n_\text{test}$ for some method, that method did not obtain counterfactuals for all observations.
We calculate the \textit{validity}, \textit{actionability}, and \textit{low-cost} metrics defined above and present the average and standard deviations across all test observations that obtained a counterfactual.  The evaluation of the \textit{data manifold closeness} is treated in Section \ref{sec:quality}.

\begin{table}[ht!]
\caption{Experiment 1: Average and standard deviation (in parentheses) of performance metrics for counterfactuals generated with MCCE and the competing methods.  
The \textbf{bold} values indicate the best values per metric and data set. `time(s) all' indicates the time (seconds) it takes to generate counterfactuals for $n_\text{test}$ observations.
}
\label{tab:results}
\centering
\begin{tabular}{lrrrrrr}
\hline
\multicolumn{7}{c}{Data set: Adult, $n_\text{test} = 1000$, $K = 1000$} \\
\hline
\hline
Method & $L_0\downarrow$ & $L_1\downarrow$ & violation$\downarrow$ & success$\uparrow$ & $N_{\text{CE}}\uparrow$ & time(s)$\downarrow$ \\
\hline
 C-CHVAE &  7.76 (1.02) &   3.13 (1.10)  &  \bf{0.00} (0.00) &     \bf{1.00} &  \bf{1000} &           86.29 \\
 CEM-VAE &  6.92 (2.06) &  3.18 (1.65)  &  1.38 (0.59) &       0.50 &  \bf{1000}  &          646.29 \\
    CLUE &   13.00 (0.00) &  7.83 (0.31) &  1.36 (0.48) &      \bf{1.00} &  \bf{1000}  &         2754.35 \\
   CRUDS &  7.87 (1.08) &   3.90 (1.11) &  \bf{0.00} (0.00) &   \bf{1.00}   &  \bf{1000}  &        10017.76 \\
    FACE &  7.03 (1.58) &  3.27 (1.54)  &  1.40 (0.52) &      \bf{1.00} &  \bf{1000}  &         4151.14 \\
  REViSE &  5.54 (0.96) &  1.13 (0.83)  &  \bf{0.00} (0.00) &   \bf{1.00}  &  \bf{1000}  &         8400.42 \\
    MCCE &  \bf{2.70} (0.73) &  \bf{0.56} (0.45) &  \bf{0.00} (0.00) &  \bf{1.00}  &  \bf{1000} &  \bf{15.00} \\
    \hline
\multicolumn{7}{c}{Data set: Give Me Some Credit, $n_\text{test} = 1000$, $K = 1000$} \\
\hline
\hline
Method & $L_0\downarrow$ & $L_1\downarrow$ & violation$\downarrow$ & success$\uparrow$ & $N_{\text{CE}}\uparrow$ & time(s)$\downarrow$ \\
\hline
  C-CHVAE &  8.98 (0.15) &  0.95 (0.29) &   \bf{0.00} (0.00) &     \bf{1.00} &  \bf{1000} &           94.38 \\
 CEM-VAE &  8.62 (1.08) &  1.61 (0.57) &    0.96 (0.19) &    0.93 &  \bf{1000}  &          685.88 \\
    CLUE &  10.00 (0.04) &  1.41 (0.32) &   1.00 (0.03) &     \bf{1.00} &  \bf{1000}  &         2533.57 \\
   CRUDS &   9.00 (0.00) &  1.68 (0.36) &   \bf{0.00} (0.00) &     \bf{1.00} &  \bf{1000}  &         8137.49 \\
    FACE &  8.53 (1.08) &  1.65 (0.53)  &   0.98 (0.16) &     \bf{1.00} &  \bf{1000}  &        13986.24 \\
  REViSE &  8.36 (1.06) &   0.70 (0.27) &    \bf{0.00} (0.00) &     \bf{1.00} &  \bf{1000} &         5722.87 \\
    MCCE &  \bf{4.52} (0.97) &  \bf{0.61} (0.32) &   \bf{0.00} (0.00) &   \bf{1.00} &  \bf{1000} &    \bf{19.58} \\
    \hline
\hline
\multicolumn{7}{c}{Data set: German Credit, $n_\text{test} = 200$, $K = 1000$} \\
\hline
\hline
Method & $L_0\downarrow$ & $L_1\downarrow$ & violation$\downarrow$ & success$\uparrow$ & $N_{\text{CE}}\uparrow$ & time(s)$\downarrow$ \\
\hline
 C-CHVAE &  10.62 (1.23) &  4.79 (1.31) &  \bf{0.00} (0.00) &     \bf{1.00} &   \bf{200} &           12.88 \\
 CEM-VAE &  10.16 (1.89) &  6.97 (1.93) &  1.44 (0.53) &    0.67 &    63 &          147.86 \\
    CLUE &  20.09 (0.00) &  13.4 (0.53) &  1.55 (0.50) &     \bf{1.00} &   \bf{200} &          719.94 \\
   CRUDS &  13.01 (1.29) &  9.22 (1.28) &  \bf{0.00} (0.00) & \bf{1.00} &   \bf{200} &         2453.17 \\
    FACE &   9.75 (2.10) &  6.68 (1.97) &  1.50 (0.55) &    \bf{1.00} &   \bf{200} &           73.68 \\
  REViSE &   7.56 (1.56) &  2.28 (1.29) &  \bf{0.00} (0.00) &    \bf{1.00} &   172 &         2021.15 \\
    MCCE &   \bf{3.62} (0.93) &  \bf{1.90} (0.76) &  \bf{0.00} (0.00) &    \bf{1.00} &   \bf{200} &            \bf{3.71} \\
   \hline
\multicolumn{7}{c}{Data set: FICO, $n_\text{test} = 1000$, $K = 1000$} \\
\hline
\hline
Method & $L_0\downarrow$ & $L_1\downarrow$ & violation$\downarrow$ & success$\uparrow$ & $N_{\text{CE}}\uparrow$ & time(s)$\downarrow$ \\
\hline
  C-CHVAE & 21.99 (0.09) &  2.08 (0.84) &   \bf{0.00} (0.00) &   \bf{1.00} &  \bf{1000} &           16.39 \\
 CEM-VAE &  19.52 (2.46) &  3.11 (0.87) &   0.95 (0.21) &     0.31 &   478 &          734.83 \\
    CLUE &  23.00 (0.00) &  2.40 (0.56) &   1.00 (0.00) &     \bf{1.00} &     99 &         3523.77 \\
   CRUDS &  22.00 (0.00) &  9.32 (2.20) &   \bf{0.00} (0.00) &     \bf{1.00} &   592 &         7639.12 \\
    FACE &  19.11 (2.32) &  3.01 (0.90) &   0.98 (0.16) &     \bf{1.00} &  \bf{1000} &          428.44 \\
  REViSE &  21.71 (0.78) &  \bf{1.64} (0.54) &   \bf{0.00} (0.00) &     \bf{1.00} &   926 &         5589.01 \\
    MCCE &  \bf{12.67} (1.91) &  1.95 (0.78) &   \bf{0.00} (0.00) &     \bf{1.00} &  \bf{1000} &         \bf{15.47} \\
\hline
\end{tabular}
\end{table}

The average and standard deviations of each metric, method, and data set are presented in Table \ref{tab:results}. 
`time(s)' indicates the time (in seconds) it takes to generate counterfactuals for $n_\text{test}$ observations. Although almost all methods produce only valid counterfactuals (average success = 1).   MCCE, \texttt{C-CHVAE} and \texttt{REViSE} are the only methods that always produce actionable counterfactuals (average violation = 0), and MCCE and \texttt{C-CHVAE} are the only ones that obtain counterfactuals for all test observations in all data sets. MCCE produces counterfactuals with 2-6 fewer feature changes than the best performing competing methods ($L_0$), and for three of the four data sets, MCCE produces counterfactuals closest to the corresponding test observations on average ($L_1$). For the FICO data set, 
the counterfactuals produced by \texttt{REViSE} are of slightly lower cost.  Finally, MCCE is faster than all the other methods for all four data sets.

To get some insight into the counterfactuals generated by the different methods, we manually analyzed the Adult counterfactuals to see whether we could detect patterns across methods. 
One example of the counterfactuals generated for the same random observation is shown in Table \ref{tab:adult_example}. 
We show that \texttt{CLUE} and \texttt{CRUDS} produce counterfactuals that are far outside acceptable bounds and \texttt{CEM-VAE} produces counterfactuals that are not valid. The remaining methods tend to change the same three features (namely \texttt{fnlwgt}, \texttt{education-number}, and \texttt{capital-gain}). However, they don't always change them equally. For example, \texttt{education-number} is increased by MCCE for all test observations, while it is increased for only 50\% of test observations by \texttt{C-CHVAE} and \texttt{REViSE}. In addition, the methods differ in how much they change the features.
\texttt{C-CHVAE} changes nearly all features by a small amount while \texttt{FACE} changes fewer features but by larger amounts. This analysis shows that, unsurprisingly, there are systematic differences between the methods.

\begin{sidewaystable}
\caption{Experiment 1: The counterfactuals generated for the same random factual of Adult.
The \textbf{bold} values indicate the values that are the same as the original factual values.
}
\label{tab:adult_example}
\centering
\begin{tabular}{lccccccccccccc}
  \hline
Method  & \multicolumn{1}{p{0.1cm}}{\centering Age} & 
\multicolumn{1}{p{0.2cm}}{\centering FNLWGT} &
\multicolumn{1}{p{0.2cm}}{\centering Edc.} & 
\multicolumn{1}{p{0.2cm}}{\centering Gain} & 
\multicolumn{1}{p{0.2cm}}{\centering Loss} &  
\multicolumn{1}{p{0.2cm}}{\centering Hr.} & 
\multicolumn{1}{p{0.3cm}}{\centering MS} & 
\multicolumn{1}{p{0.2cm}}{\centering Co.} &
\multicolumn{1}{p{0.3cm}}{\centering Occ.} &
\multicolumn{1}{p{0.2cm}}{\centering Race} &
\multicolumn{1}{p{0.2cm}}{\centering Rel.} &
\multicolumn{1}{p{0.2cm}}{\centering Sex} &
\multicolumn{1}{p{0.2cm}}{\centering Work} 
\\
\hline
\textbf{Original} & \textbf{20} & \textbf{\numprint{266015}} & \textbf{10} & \textbf{0} & \textbf{0} & \textbf{44} & \textbf{NM} & \textbf{US} & \textbf{O} & \textbf{NW} & \textbf{NH} & \textbf{M} & \textbf{P} \\
\hdashline[1pt/5pt]
  C-CHVAE &   \textbf{20} &  \numprint{247240} &             \textbf{10} &           652 &          1679 &              42 &              M &             \textbf{US} &          \textbf{O} &   \textbf{NW} &            H &   \textbf{M} &         \textbf{P} \\
  CEM-VAE &   23 &  \numprint{190709} &             12 &         \numprint{10296} &             \textbf{0} &              52 &             \textbf{NM} &             \textbf{US} &          \textbf{O} &    W &           \textbf{NH} &   \textbf{M} &        NP \\
     CLUE &   11 &  \numprint{398962} &             \textbf{10} &         \numprint{10725} &           -62 &              49 &             \textbf{NM} &             \textbf{US} &          \textbf{O} &    W &           \textbf{NH} &  \textbf{ M} &         \textbf{P} \\
    CRUDS &   \textbf{20} &  \numprint{138021} &             21 &          4833 &           210 &              79 &              M &             \textbf{US} &         MS &    W &            H &   M &         \textbf{P} \\
     FACE &   46 &  \numprint{220979} &             \textbf{10} &         \numprint{13550} &             \textbf{0} &              40 &             \textbf{NM} &             \textbf{US} &          \textbf{O} &   \textbf{NW} &           \textbf{NH} &   \textbf{M} &         \textbf{P} \\
   REViSE &   \textbf{20} &   \numprint{76456} &             14 &            10 &           379 &              68 &              M &             \textbf{US} &          \textbf{O} &    W &            H &   \textbf{M} &         \textbf{P} \\
     \textbf{MCCE} &   \textbf{20} &  \numprint{348148} &             \textbf{10} &         \numprint{34095} &             \textbf{0} &              48 &             \textbf{NM} &             \textbf{US} &          \textbf{O} &   \textbf{NW} &           \textbf{NH} &   \textbf{M} &         \textbf{P} \\
\hline
\end{tabular}
\end{sidewaystable}

\subsection{Experiment 2: How does MCCE's performance vary with \texorpdfstring{$K$}?}\label{sec:increase_K}
Compared to all competing methods other than \texttt{FACE}, MCCE has only one tuning parameter, the
number of samples generated, $K$ (see Appendix \ref{parValues}). In this section, we study how MCCE's performance varies with $K$. We repeat the analysis for the Adult data set in Experiment 1 varying $K$ in $[10, 50, 100, 1000, \numprint{10000}, \numprint{25000}]$ and examine both MCCE's metric averages and run times.
We show the results in Table \ref{tab:timing_increase_K}.  `model(s)', `gener(s)', and `post-proc(s)' indicate the time (in seconds) it takes for MCCE's modeling, generating, and post-processing steps to run, respectively. `total(s)' is the total time. 

We see that a $K$ as low as 10 generates valid counterfactuals for 99.9\% of test observations and gives results of $L_0$ and $L_1$ that are better than all competing methods. Although the metrics become better with larger $K$, conditioning on the decision eliminates the risk of not sampling any valid counterfactuals.
Conditioning on the immutable features also ensures the samples are actionable and relatively close to the factual. 

\begin{sidewaystable}
    \caption{Experiment 2: Average and standard deviation (in parentheses) of performance metrics for counterfactuals generated with MCCE as we vary $K$. 	
    }
    \label{tab:timing_increase_K}
    \centering
    \begin{tabular}{rcccccrrrr}
    \hline
    \multicolumn{10}{c}{Data set: Adult, $n_\text{test} = 1000$, $K$ varies} \\
    \hline
    \hline
    $K$ & $L_0\downarrow$ & $L_1\downarrow$ & violation$\downarrow$ & $N_{\text{CE}}\uparrow$   & model(s)$\downarrow$  & gener(s)$\downarrow$ & post-proc(s)$\downarrow$ & total(s)$\downarrow$ \\
    \hline
      10 &   3.98 (1.1) &  1.59 (0.94) &   0 (0.0) &            999 &            3.8 &                0.38 &                   5.56 &            9.74 \\
    50 &  3.39 (0.94) &   1.1 (0.77) &   0 (0.0) &             1000 &            3.8 &                0.58 &                   5.61 &            9.99 \\
   100 &   3.18 (0.9) &  0.93 (0.69) &   0 (0.0) &             1000 &            3.8 &                0.76 &                   5.72 &           10.28 \\
  1000 &   2.7 (0.73) &  0.56 (0.45) &   0 (0.0) &             1000 &            3.8 &                4.39 &                   7.23 &           15.43 \\
  5000 &  2.51 (0.66) &  0.45 (0.39) &   0 (0.0) &             1000 &            3.8 &               21.29 &                  12.34 &           37.43 \\
 \numprint{10000} &  2.43 (0.64) &  0.43 (0.37) &   0 (0.0) &  1000 &            3.8 &               42.61 &                  17.71 &           64.11 \\
 \numprint{25000} &  2.39 (0.64) &  0.41 (0.36) &   0 (0.0) &  1000 &            3.8 &              106.93 &                  29.50 &          140.23 \\

    \hline
    \end{tabular}
\end{sidewaystable}

\begin{figure}[ht]
    \centering
    \includegraphics[width=1\linewidth]{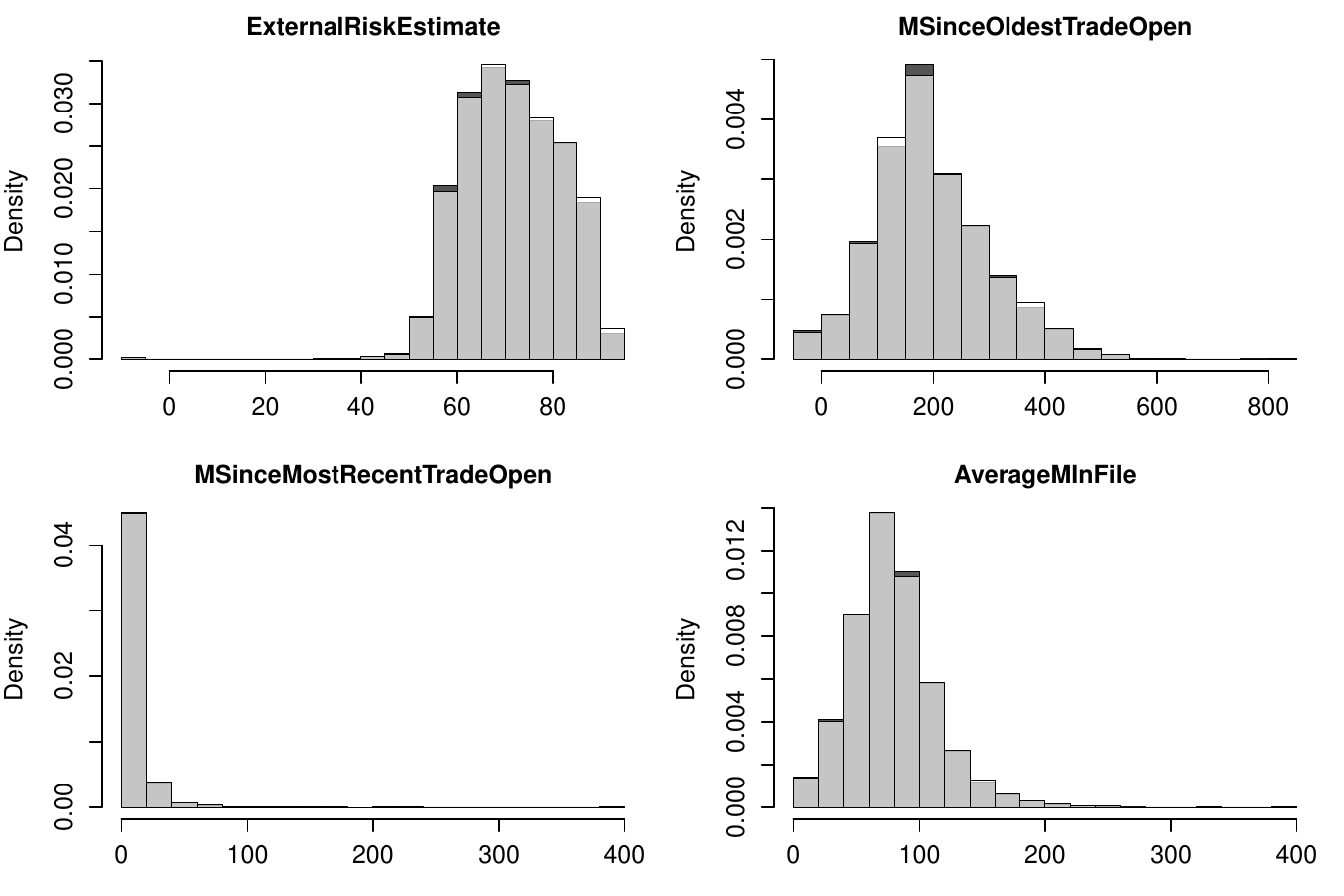} 
    \caption{Histograms for three of the variables in the generated data set (white) with the histograms for the real data superimposed (dark grey).
     Where the histograms overlap, the blend of of white and dark grey gives a light grey color.}
    \label{fig:histograms}
\end{figure}

\subsection{Experiment 3: Are MCCE counterfactuals on-manifold?} \label{sec:quality}

\begin{figure}[ht]
    \centering
    \includegraphics[width=.6\linewidth]{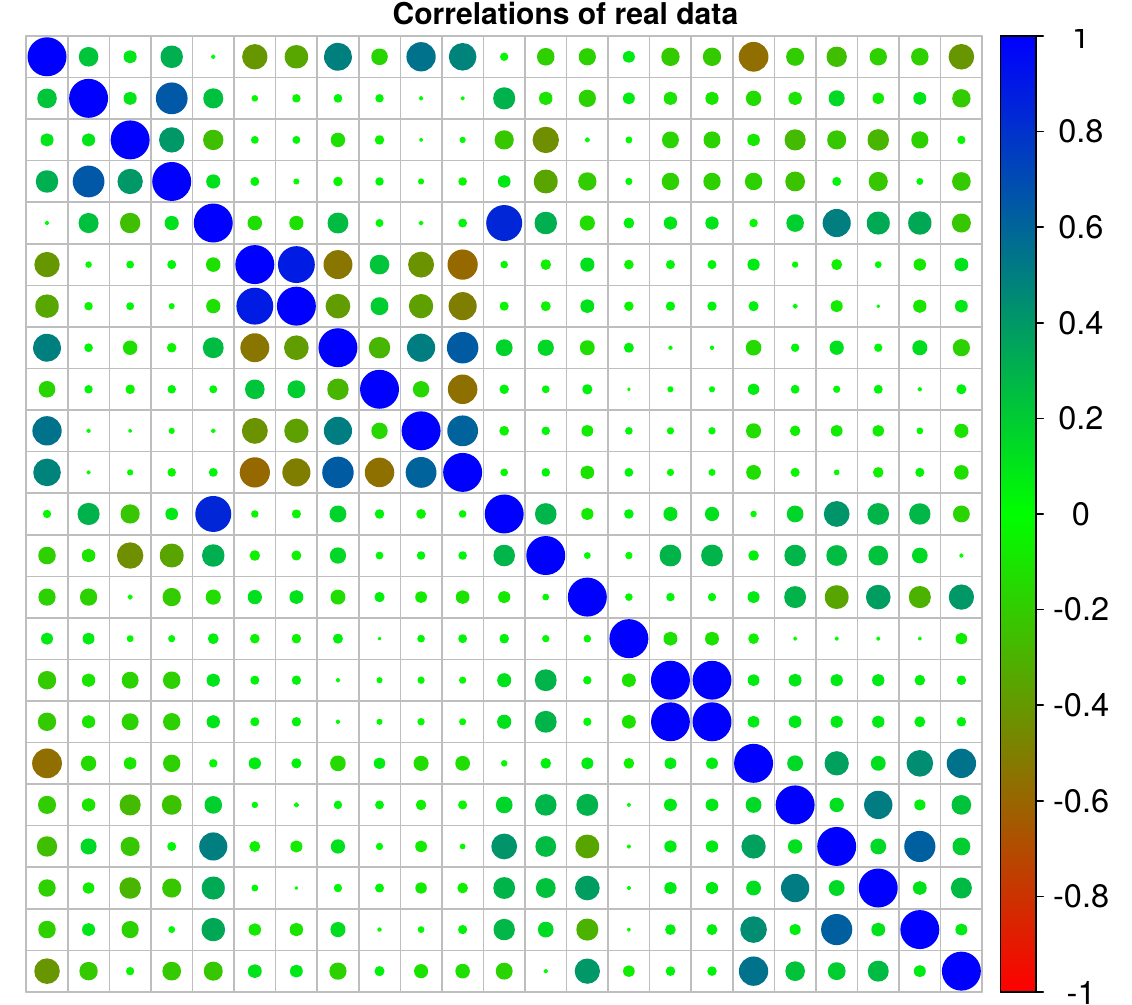} 
    \caption{Correlation matrix for the FICO data set. The areas of the circles are proportional to the absolute value of the corresponding correlation coefficients.}
    \label{fig:corMatReal}
\end{figure}

\begin{figure}[ht]
    \centering
    \includegraphics[width=.6\linewidth]{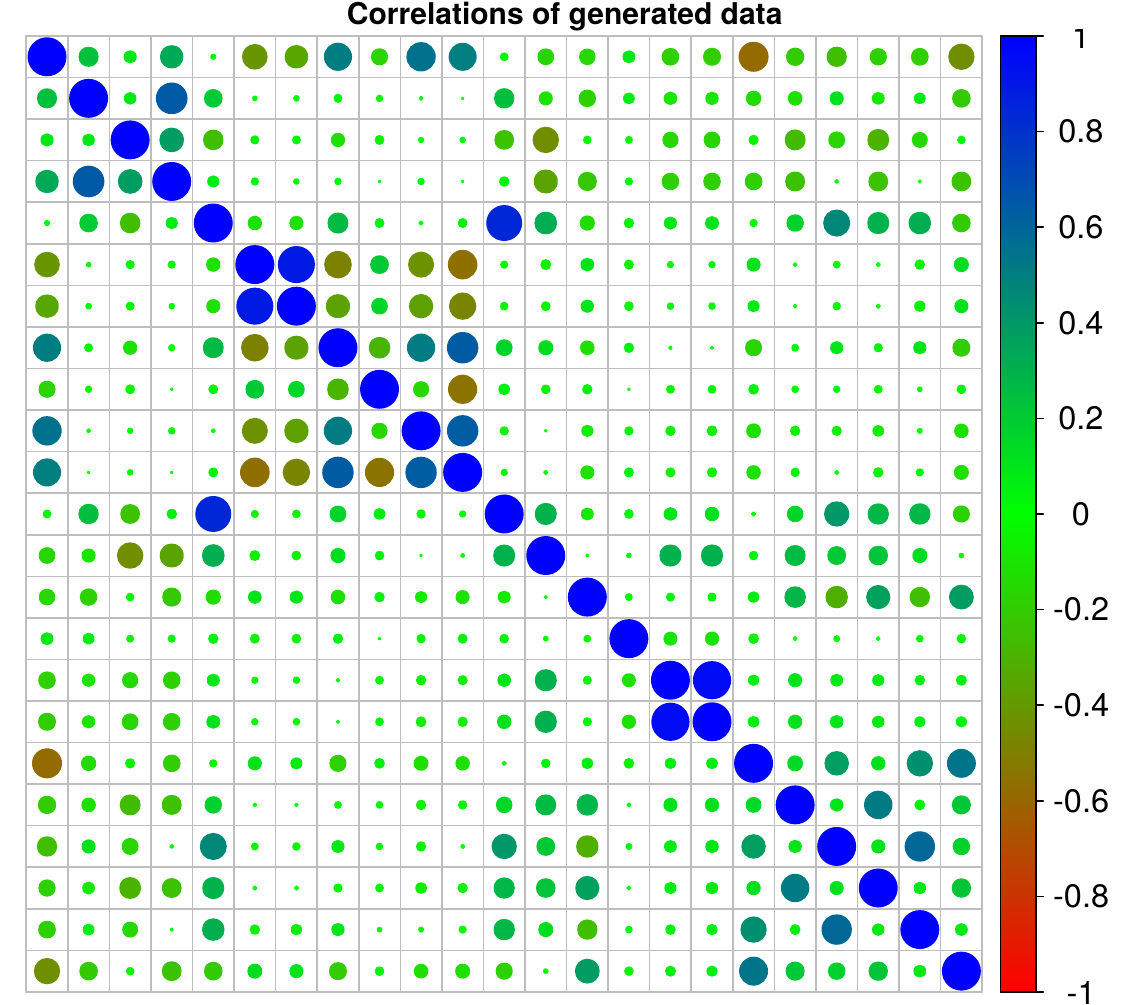} 
    \caption{Correlation matrix for the data generated from the autoregressive generative model ($K=\numprint{10000}$). The areas of the circles are proportional to the absolute value of the corresponding correlation coefficients.}
    \label{fig:corMatGen}
\end{figure}

\begin{figure}[ht]
    \centering
    \includegraphics[width=.6\linewidth]{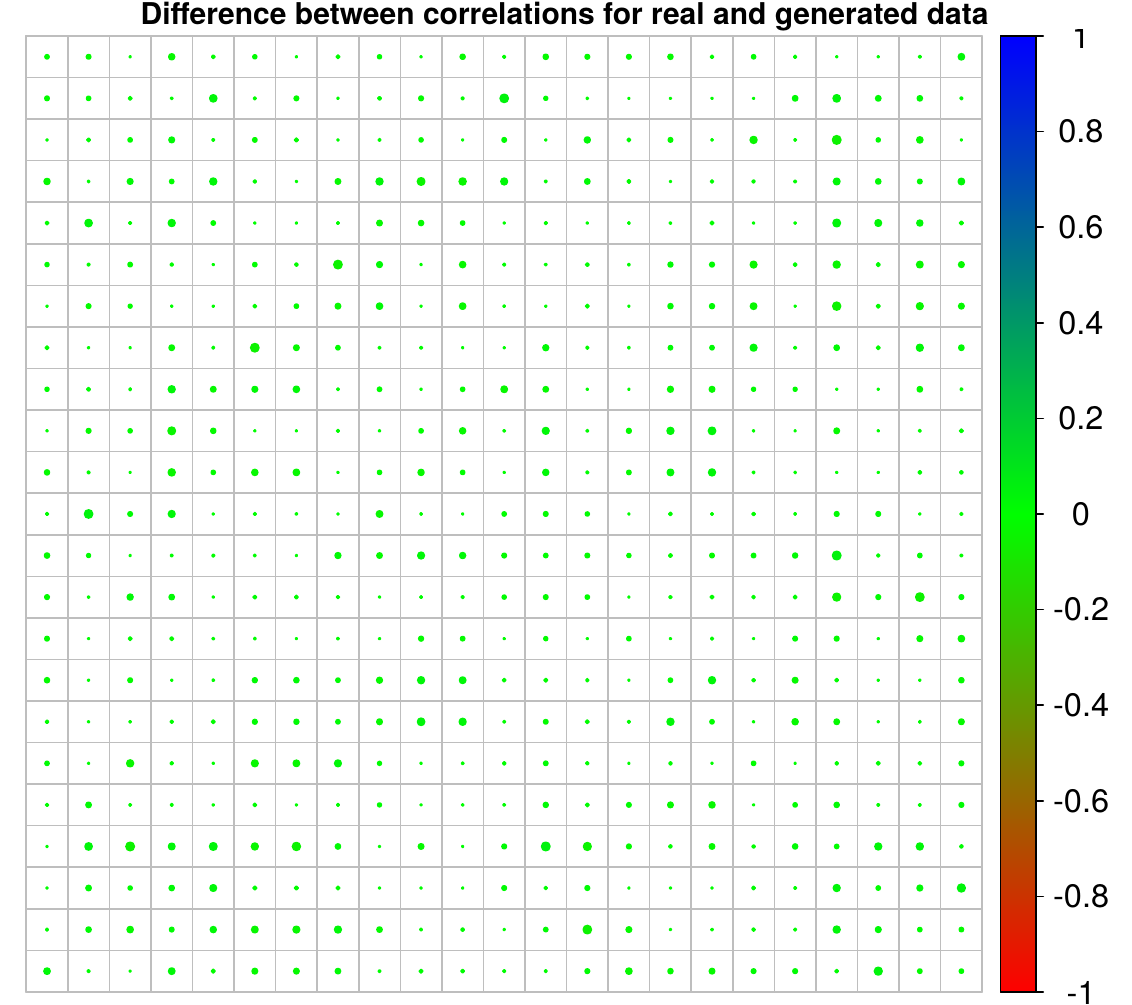} 
    \caption{Difference between the correlation matrices in Figures \ref{fig:corMatReal} and \ref{fig:corMatGen}. The areas of the circles are proportional to the absolute value of the corresponding correlation coefficients.}
    \label{fig:corMatDiff}
\end{figure}

According to \cite{guidotti2022counterfactual}, a plausible counterfactual is `realistic' if it is `similar' to the known data set and adheres to observed correlations among the features. In this section, we study the characteristics of the data generated in step 2 of the MCCE method for the FICO data set, using $K=\numprint{10000}$. Figure \ref{fig:histograms} shows histograms for four of the variables in this data set (white) with histograms of the real data superimposed (dark grey). Where the histograms overlap, the blend of white and dark grey gives a light grey color. As can be seen from the figure, the generated marginal distributions for both the continuous and categorical variables are very close to the data distributions. Appendix \ref{sec:quality2} shows that the figures for the rest of
the variables are very similar.
In Figures \ref{fig:corMatReal}-\ref{fig:corMatDiff} we show the correlation matrices for the real and generated data, as well as the difference between the correlation matrices. It is evident that even the correlations are very well captured by the generative model.  

Another way of checking whether the generated data can easily be told apart from the original data is to compute the discriminator measure described by \cite{borisov23}. To do so, we fit a Random Forest model on a mix of generated training samples and samples from the real training data set, where the response indicates whether the observation comes from the generated or real data (when performing this experiment, the real response is discarded). 
Then, we use this model to predict the probability of belonging to the real data on a set of new samples and compute the model's out-of-sample accuracy using a threshold of 0.5. 
If the discriminator cannot differentiate between the generated and real samples, the model's accuracy will be around 50\%. In our case, the mean accuracy over five trials with different random seeds for the Adult, GMC, FICO, and German Gredit data sets are 62.8\%, 56.5\%, 59.1\%, and 55.7\%, respectively. Although these numbers are slightly higher than 50\%, the numbers for Adult and FICO are lower than the lowest accuracy reported for the same datasets in \cite{borisov23}\footnote{In \cite{borisov23} the FICO dataset is referred to as the HELOC dataset.}.

These experiments show that it is difficult to differentiate between counterfactuals generated by MCCE and the underlying data set.

\subsection{Experiment 4: Is MCCE scalable?}\label{sec:limitations}

For explanability methods to be applicable in a wide range of practical situations, it is crucial that they scale well computationally. 
We now study the scalability of the different steps of the MCCE procedure. 
The first subsection provides actual measurements of the computation time for the three MCCE steps as we vary the different quantities, using a controlled simulation setup. 
The second subsection provides the corresponding theoretical computational complexity in big-$O$ notation.

\subsubsection{Computational complexity in practice}

Let the $p$-dimensional feature vector $\bm{X}$ be simulated from a zero-mean Gaussian distribution with covariance Cov$(X_j,X_k)=0.5$ and Var$(X_j)=1$ for all $j \neq k$. In this simulated setup, we explain simple linear models of the form $f(\bm{x}) = \sum_{j=1}^p \beta_j x_j$, where $\beta_j=1$ for all $j$.

We generate counterfactual explanations for predictions of this model as we vary the following quantities:
$p$,
$n_{\text{train}}$,
$n_{\text{test}}$, and
$K$.
For each simulated model, we set the decision interval for an acceptable decision to $c = (0,\infty)$, which on average covers half of the observations regardless of the dimension $p$. We always only explain test observations that are originally outside of this interval.
Moreover, we do not fix any of the features in the explanations, i.e.,~all features are mutable ($q=p$).

As an initial broad overview, Table \ref{tab:simres} shows the computation times for the three steps of MCCE for all combinations of
$p = (5,30)$,
$n_{\text{train}} = (1000, \numprint{10000})$,
$n_{\text{test}} = (1,50)$, and
$K = (\numprint{10000}, \numprint{100000})$. 
The displayed computation times represent the mean of 10 repeated computations.

\begin{table}[ht]
\caption{Overview of the computational times (in seconds) of the three steps of MCCE when varying $n_{\text{test}}$,  $n_{\text{train}}$, $p$, and $K$.
The displayed computation times represent the mean of 10 repeated computations.
}
\label{tab:simres}
\centering
\begin{tabular}{rrrr|rrrr}
  \hline
$n_{\text{test}}$ & $n_{\text{train}}$ & $p$ & $K$ & model(s)$\downarrow$ & gener(s)$\downarrow$ & post-proc(s)$\downarrow$ & total(s)$\downarrow$  \\ 
  \hline
  1 & 1000 &   5 & \numprint{10000} & 0.04 & 0.11 & 0.09 & 0.24 \\ 
    1 & 1000 &   5 & \numprint{100000} & 0.04 & 0.75 & 0.12 & 0.91 \\ 
    1 & 1000 &  30 & \numprint{10000} & 1.25 & 0.83 & 0.09 & 2.17 \\ 
    1 & 1000 &  30 & \numprint{100000} & 1.25 & 5.10 & 0.15 & 6.50 \\ 
    1 & \numprint{10000} &   5 & \numprint{10000} & 0.48 & 0.07 & 0.09 & 0.64 \\ 
    1 & \numprint{10000} &   5 & \numprint{100000} & 0.48 & 0.46 & 0.12 & 1.06 \\ 
    1 & \numprint{10000} &  30 & \numprint{10000} & 12.05 & 0.95 & 0.10 & 13.10 \\ 
    1 & \numprint{10000} &  30 & \numprint{100000} & 12.05 & 5.39 & 0.15 & 17.59 \\ 
   50 & 1000 &   5 & \numprint{10000} & 0.05 & 3.70 & 0.99 & 4.74 \\ 
   50 & 1000 &   5 & \numprint{100000} & 0.04 & 38.26 & 3.36 & 41.66 \\ 
   50 & 1000 &  30 & \numprint{10000} & 1.24 & 24.89 & 1.03 & 27.16 \\ 
   50 & 1000 &  30 & \numprint{100000} & 1.24 & 264.63 & 6.21 & 272.08 \\ 
   50 & \numprint{10000} &   5 & \numprint{10000} & 0.48 & 2.07 & 0.70 & 3.25 \\ 
   50 & \numprint{10000} &   5 & \numprint{100000} & 0.49 & 25.16 & 3.01 & 28.66 \\ 
   50 & \numprint{10000} &  30 & \numprint{10000} & 12.06 & 24.87 & 1.04 & 37.97 \\ 
   50 & \numprint{10000} &  30 & \numprint{100000} & 12.00 & 268.76 & 6.03 & 286.79 \\ 
   \hline
\end{tabular}
\end{table}

Altering all quantities affects the total computation times. 
Even for as many as $p=30$ features, performing the modeling with $n_{\text{train}} = \numprint{10000}$ only takes 12 seconds. The dimension $p$ does, however, play a crucial role for both the modeling time and the generation time. 
Further, the number of samples generated per test observation, $K$, and the number of test observations $n_{\text{test}}$, naturally affects the generation time significantly. The post-processing time is quite limited even for $n_{\text{test}}=50,\ p=30$ and $K=\numprint{100000}$. That is, for the model and quantity sizes used here, the generation time is the driving factor.

\begin{figure}[ht]
    \centering
    \includegraphics[width=\linewidth]{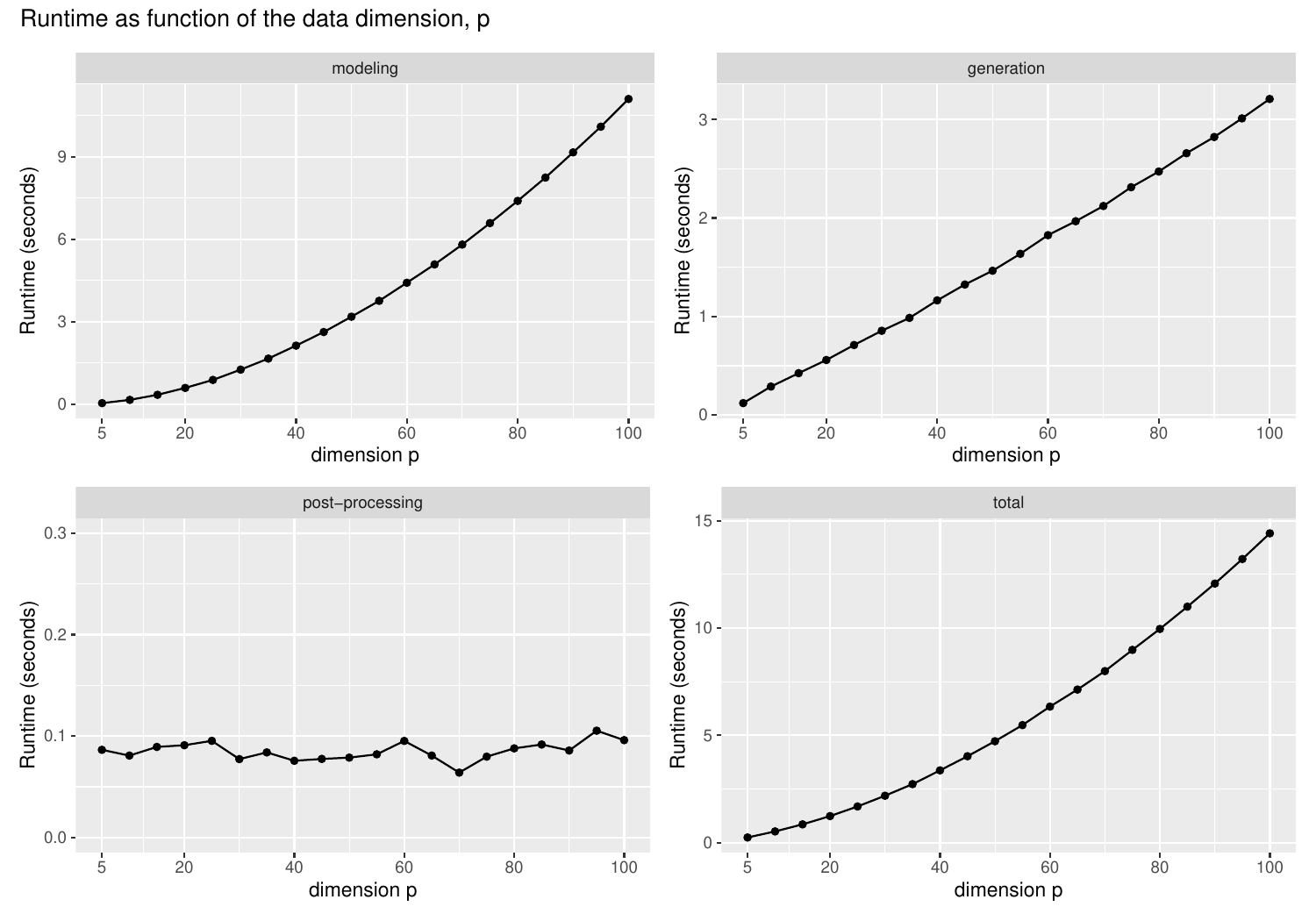} 
    \caption{Runtime as function of dimension $p$. The other quantities are fixed to $n_{\text{test}}=1,\ n_{\text{train}}= 1000,\ K = \numprint{10000}$.}
    \label{fig:runtime_p}
\end{figure}

\begin{figure}[ht]
    \centering
    \includegraphics[width=\linewidth]{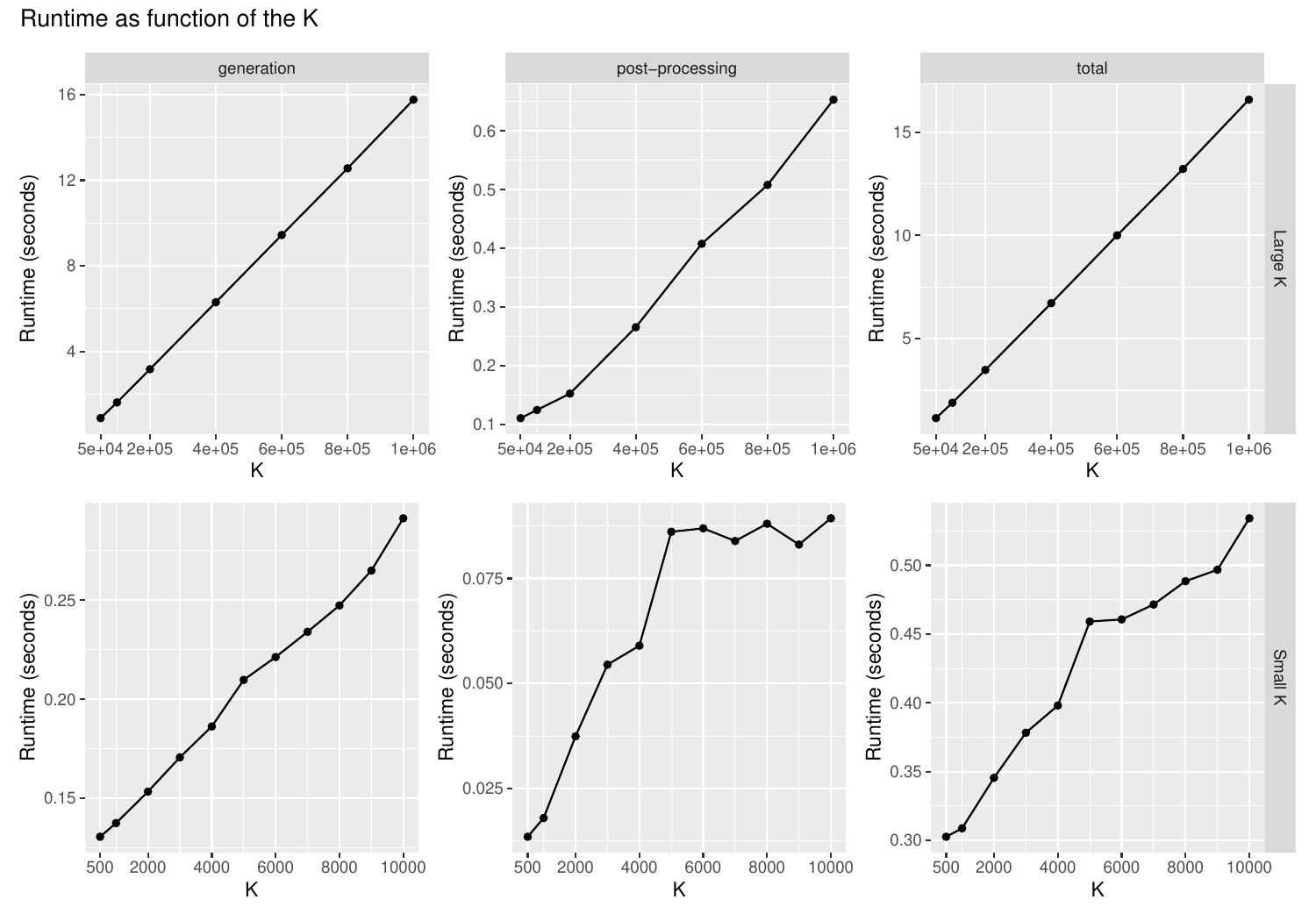} 
    \caption{Runtime as function of the number of generated samples $K$. The other quantities are fixed to $n_{\text{test}}=1,\ p=10,\ n_{\text{train}} = 1000$. The upper panels show $K > 10^4$, and lower panels show $K\leq 10^4$.
    The modeling time is obviously unaffected by $K$ and is therefore not displayed.}
    \label{fig:runtime_K}
\end{figure}

\begin{figure}[ht]
    \centering
    \includegraphics[width=\linewidth]{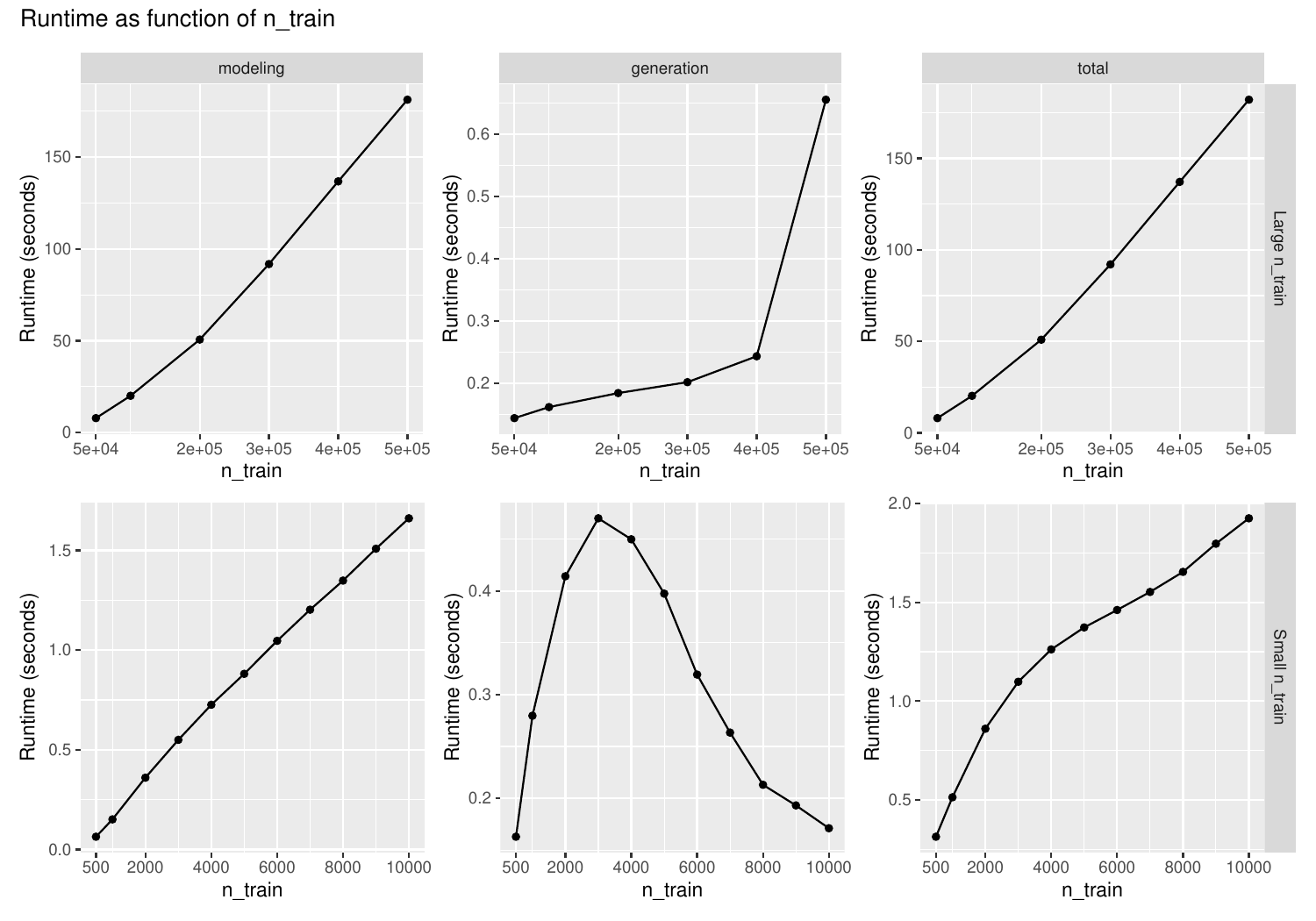} 
    \caption{Runtime as function of number of training samples $n_{\text{train}}$.  The other quantities are fixed to $n_{\text{test}}=1,\ p = 10,\ K = \numprint{10000}$.
    The upper panels shows $n_{\text{train}} > 10^4$ and the lower panels shows $n_{\text{train}}\leq 10^4$.
    The post-processing time is not affected by $n_{\text{train}}$ and is therefore not displayed.}
    \label{fig:runtime_ntrain}
\end{figure}

\begin{figure}[ht]
    \centering
    \includegraphics[width=\linewidth]{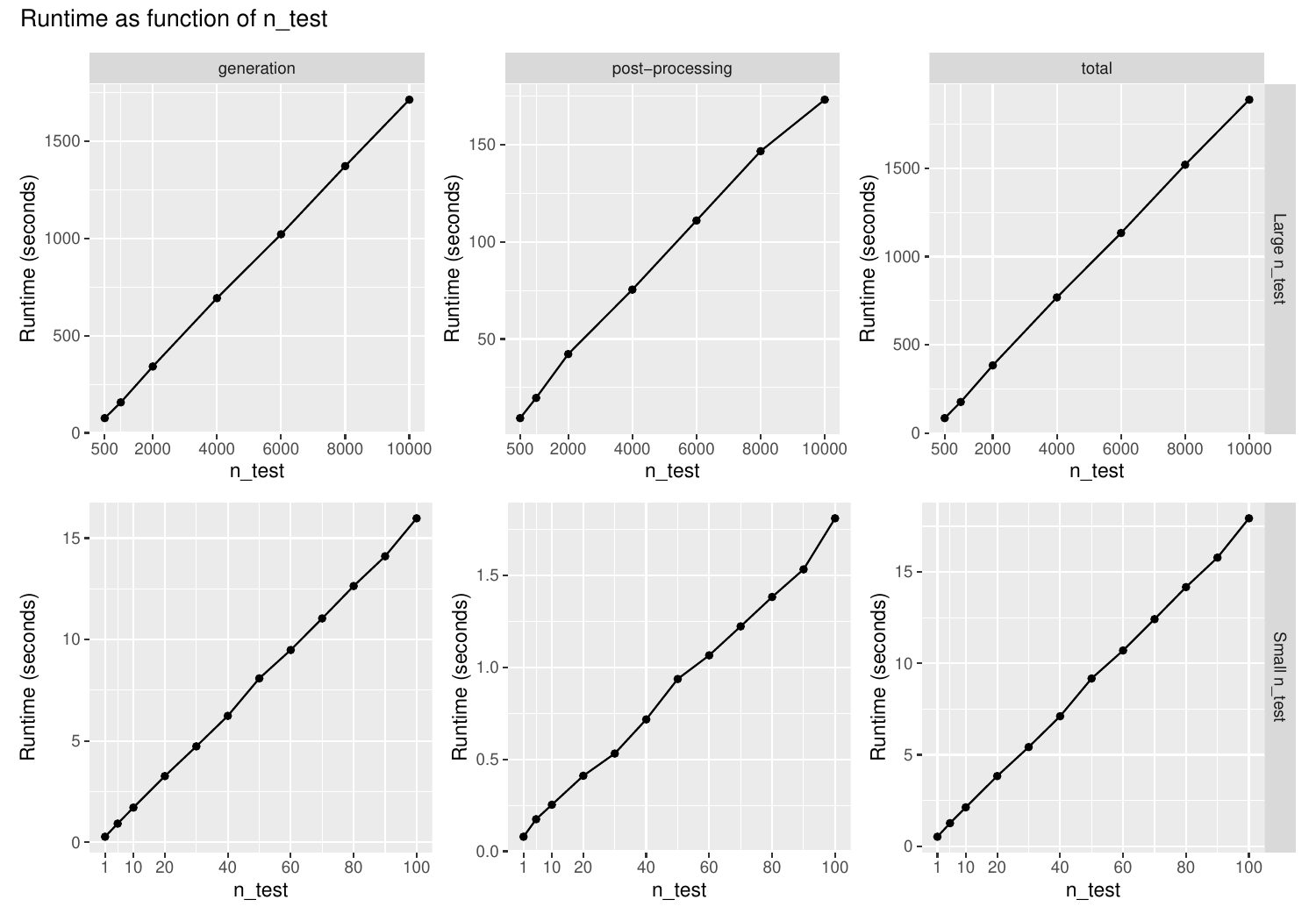} 
    \caption{Runtime as function of number of test observations $n_{\text{test}}$.  The other quantities are fixed to $n_{\text{train}}= 1000,\ p = 10,\ K = \numprint{10000}$.
        The upper panels shows $n_{\text{test}} > 10^4$, and upper panels shows $n_{\text{test}}\leq 10^4$.
        The modeling time is clearly not affected by $n_{\text{test}}$ and is therefore not displayed.}
    \label{fig:runtime_ntest}
\end{figure}

In Figures \ref{fig:runtime_p}-\ref{fig:runtime_ntest} we take a more detailed look at how each of the four quantities affects the runtime of the different steps when altering one at a time and keeping the others fixed. When the quantities are not altered, we fix them at the following values: $n_{\text{test}} = 1,\ n_{\text{train}} = 1000,\ p = 10,\ K = \numprint{10000}$.

From Figure \ref{fig:runtime_p}, we see that increasing the dimension $p$, increases the modeling time faster-than-linearly, the generation time linearly, and the post-processing time not at all. The higher rate of the modeling time indicates that for very high dimensions ($> 100$), the modeling time may be the driving factor.

Figure \ref{fig:runtime_K} shows that the generation time scales linearly at a significant rate in terms of $K$.
The picture is not as clear for the post-processing time, but this is less important, as the actual run times are rather small for this component. The modeling time is obviously unaffected by $K$ and is therefore not displayed.

Figure \ref{fig:runtime_ntrain} shows that the generation time scales approximately linearly at quite a high rate. 
The effect of $n_{\text{train}}$ on the generation time is unclear from these simulations. 
For the smallest values of $n_{\text{train}}$, this component is probably influenced by inaccuracies in the computational timing. 
The post-processing time is not affected by $n_{\text{train}}$ and is therefore not displayed.

Figure \ref{fig:runtime_ntest} shows that both the generation and post-processing times naturally scale approximately linearly in terms of $n_{\text{test}}$. The modeling time is clearly not affected by $n_{\text{test}}$ and is therefore not displayed.
For this specific setup, the total computation time seems to increase by about 0.2 seconds per extra test observation.

To showcase that these scalability results are relevant and valid in a broader context, we also run simulations with quantities mimicking the four real data sets from Section \ref{subsec:experiment1}. Table \ref{tab:datasimres} in Appendix \ref{sec:real_data_sim} displays the computation times for these simulations and shows a relatively good match between the computation times in the simulations and the real data experiments. This suggests that the scalability observed in the above simulation study is likely to extend roughly to real-world data scenarios.

\subsubsection{Theoretical computational complexity}

The simulation study above shows that MCCE scales well in all four quantities.  In this section, we discuss the theoretical computational complexity in terms of big-O notation. 
Recall that $u$ and $q$ are the number of immutable and mutable features, respectively.

According to \citet[Ch 9.7]{hastie2009elements}, the training time for the $i$-th tree (with $u+i$ features) is $O((u+i)  n_{\text{train}} \log(n_{\text{train}}))$. Since we have $q-1$ such trees and $\sum_{i=1}^{q-1}(u+i) = (q-1)(u+(q-2)/2)$, the computational complexity of step 1 is $O(q (u + q)n_{\text{train}}\log(n_{\text{train}}))$.

Generating a sample from one of the conditional distributions in step 1 requires traversing the decision tree from root to leaf, which requires going through roughly $O(\log(n_{\text{train}}))$ nodes \citep{Geron}. 
Hence, the computational complexity of step 2 is $O(n_\text{test}Kq\log(n_{\text{train}}))$.

In step 3, we compute validity, sparsity, and the Gower distance for every sample to find the best counterfactual. 
The computational complexity of the validity computation is non-trivial since it depends on the choice of prediction model. Assuming that the computation time of the prediction model increases no faster than linearly in $u+q$, the computational complexity of step 3 is $O\left(n_\text{test}\, K\, (u+q)\right)$.  

Removing smaller order terms, the total computational complexity of MCCE becomes $O\left(q(u+q) n_{\text{train}} \log(n_{\text{train}}) + n_\text{test}K(q\log(n_{\text{train}}) + u)\right)$, showing that MCCE scales very well in all quantities 
(i.e., quadratic in $q$, at rate $n_{\text{train}}\log(n_{\text{train}})$ in $n_{\text{train}}$, and linear in $u, K$, and $n_\text{test}$).

Overall, these theoretical scalability terms are in accordance with our findings from the simulations: 
For the modeling time, the quadratic increase in $q$ and log-linear increase in $n_{\text{train}}$ match the findings from our simulations.
The linear increase in $n_{\text{test}}, K$, and $q$ for the generation time is also in accordance with our simulation results. The scalability in terms of $n_{\text{train}}$ was, as mentioned, fairly uncertain, partly due to the low computation time ($< 1$ sec). Thus, it is not unreasonable that the theoretically found logarithmic rate is valid also for our implementation (with quite a small factor).
For the post-processing time, the computation times in our simulations are so small that the theoretically found linearity in $q$, $K$ and $n_{\text{test}}$ seems reasonable.

\section{Privacy}\label{privacy}
Naive counterfactual explanations can leak sensitive or personal data in a variety of ways. First, if a counterfactual explanation is an exact copy of an individual in the training data, this individual may be recovered, leading to a breach of privacy. 
Table \ref{tab:copies} presents both the number and percentage of MCCE counterfactuals with exact copies in each of the datasets discussed in this paper. Since there are at most 1.1\% of counterfactuals with copies, this type of privacy breach is rare for MCCE.

\begin{table}[ht!]
    \caption{The number and proportion of counterfactuals being copies of original training observations for each dataset.}
    \centering
    \begin{tabular}{l|r|r|r|r}
         \hline
         & Adult & GMC & FICO & German Credit\\
         \hline
         \# counterfactuals  & 1000 & 1000 & 1000 & 200\\
         \# copies    &   11 &  4 &      0 &    2\\ 
         \hline
         Proportion of copies        &    1.1\%  &    0.4\% &     0\% &    1.0\%\\
         \hline
    \end{tabular}
    \label{tab:copies}
\end{table}

However, in certain circumstances, it may be necessary for a strict assurance that the counterfactuals generated by MCCE do not match training observations. To achieve this, one can easily add a step in MCCE's post-processing step to remove all rows of $\mathbf{D}_h$ that exactly match an observation in the training data.
Although this approach may slightly diminish the quality of the counterfactuals based on other performance measures, it provides a safeguard against privacy breaches of this type.

Another privacy breach may occur due to so-called explanation linkage attacks \citep{goethals2022privacy} or membership inference attacks. A linkage attack is an attempt to re-identify individuals in a dataset by combining the generated data with background information.
A membership inference attack attempts to re-identify individuals by exploiting the fact that their counterfactual explanation may reveal whether they were used in training the model \citep{pawelczyk2023privacy}. 
Even though MCCE's explanations are rarely observed in the underlying data, it is an open question as to whether MCCE counterfactuals are vulnerable to explanation linkage or membership inference attacks. One way to alleviate these potential privacy problems is to incorporate differential privacy \citep{dwork2006differential}.
In the case of MCCE, incorporating differential privacy could entail introducing noise in its data generating step (see e.g.,~\citet{mahiou2022dpart} or \citet{nowok2016synthpop}). Further work is required to investigate the efficiency and practical considerations of such an approach.

\section{Conclusion}\label{sec:conclusion}

To the best of our knowledge, MCCE is the first counterfactual method that models both the underlying data distribution and the decision. As shown in our experiments, this is a powerful setup because even without the post-processing step, MCCE generates samples that are almost always valid and actionable.  
In addition, modeling the data with an autoregressive generating model produces counterfactuals with lower costs than all the competing methods.
Finally, as opposed to other on-manifold methods that typically use variational autoencoders and strict prediction model and data requirements, MCCE handles any type of prediction model and categorical features with more than two levels.

\subsection{Future steps}
MCCE's three steps are flexible and can be improved independently of each other. 
We will discuss possible changes to each step below. 
In MCCE's first step, the decision trees can easily be replaced with another generation method like conditional generative adversarial nets (CGAN) \citep{mirza2014conditional} or conditional tabular GANs (CTGAN) \citep{xu2019modeling}. Note, however, that both CGANs and GANs are restricted to gradient-based classifiers. If formal privacy guarantees are required, differential privacy can be built into MCCE's data-generating process with little effort.

In MCCE's second step (the autoregressive generative model), the visit sequence has to be chosen. In the experiments conducted in this paper, the visit sequence corresponded to the order of variables in the original data sets. One might get even better correspondence between the generated and original data if the visit sequence for example is chosen based on the correlation between the variables, modeling the most correlated variables first. 
This would be an interesting and relevant topic for further work.

Finally, MCCE's third step can easily be adjusted to handle other performance metrics. 
Since the Gower distance penalizes changes to continuous features less than categorical features, one might choose to replace it with another type of distance function like the Heterogeneous or Interpolated Value Difference Metrics \citep{wilson1997improved}.
If, for some reason, the number of features being changed is irrelevant for a specific application, the initial filtering with the $L_0$ norm may also be dropped.

We have developed MCCE for tabular data. One avenue for further work is to adapt MCCE to non-tabular data. As long as the prediction model is used for a binary (or multiclass) decision, a prerequisite for counterfactual explanations, it should be possible to adapt MCCE's three steps to non-tabular data, such as time series, text, or images. One challenge might be that the number of samples needed to generate explanations with low enough cost becomes higher with more complex data.

In this paper, the aim was to generate one counterfactual for each individual. However, as mentioned, MCCE can easily be extended to return the $k \in K$ best counterfactuals if this is desired.
To enhance the practical relevance of a set of multiple counterfactuals, it might be desirable for the generated counterfactuals to also exhibit \textit{diversity} \citep{mothilal2020explaining}. Finding the satisfactory balance between performance scores for single counterfactuals and their diversity remains a subject for future research.

\backmatter


\bmhead{Acknowledgments}

This work was supported by the Norwegian Research Council grant 237718 (BigInsight). 
This paper is supported by the European Union’s HORIZON Research and Innovation Programme under grant agreement No 101120657, project ENFIELD (European Lighthouse to Manifest Trustworthy and Green AI). 

\bmhead{Authors' contributions}

AR: Methodology, Software, Validation, Data Creation, Original Draft, Writing, Review \& Editing, Visualization.
MJ: Conceptualization, Methodology, Software, Validation, Original Draft, Writing, Review \& Editing, Visualization.
KA: Conceptualization, Methodology, Validation, Original Draft, Writing, Review \& Editing, Visualization.
AL: Conceptualization, Methodology, Writing, Original Draft, Review \& Editing, Visualization.

\bmhead{Availability of data and materials}

The Adult, FICO and German Credit Data Sets can be downloaded from 
\url{https://github.com/riccotti/Scamander/blob/main/dataset} 
and the Give Me Some Credit dataset from 
\url{https://www.kaggle.com/c/GiveMeSomeCredit/data}.

\section*{Declarations}

\bmhead{Conflict of interest} The authors declare that they have no conflicts of interest.

\bmhead{Code availability}
The Python code used in this paper is open source, and can be downloaded at \url{https://github.com/NorskRegnesentral/mccepy}. A similar R package is available at \url{https://github.com/NorskRegnesentral/mcceR}.

\bmhead{Ethics approval} Not applicable.

\bmhead{Consent to participate} Not applicable.

\bmhead{Consent for publication} The authors declare that they provide consent for publication.

\clearpage 

\begin{appendices}

\section{Further experiments}

\subsection{How is MCCE's performance when the categorical features are not binarized?} \label{non-binarised}

Since none of the on-manifold methods investigated in the main paper handle categorical features with more than two levels, we had to restrict the models in the method comparisons to models with only two levels. The MCCE method can, however, handle models with categorical features with an arbitrary number of levels. To exemplify this, we here provide performance results for the MCCE applied to the Adult data set \textit{without} binarizing the seven categorical features. The results are shown in 
Table \ref{tab:results_non_binary}.

Since the explanation is carried out on different data/model, the performance scores are not directly comparable. Note, however, that the costs are slightly higher for these counterfactuals, and that the computation time also is higher. Both of these effects are expected since the data are much richer -- one of the categorical features (country) has, for instance, 41 different levels.
Changes in the categorical features are then more likely. Training the decision trees is also more time-consuming as many more splits need to be considered.

\begin{table}[ht]
    \caption{Experiment 5: Average and standard deviation (in parentheses) of performance metrics for counterfactuals generated by MCCE when the categorical features are \textbf{not binarized}.
    }
\label{tab:results_non_binary}
\centering
\begin{tabular}{ccccccc}
\hline
\multicolumn{7}{c}{Data set: Adult, $n_\text{test} = 1000$, $K = 1000$} \\
\hline
\hline
 $L_0\downarrow$ & $L_1\downarrow$  & violation$\downarrow$ & success$\uparrow$ & $N_{\text{CE}}\uparrow$ & time(s)$\downarrow$ \\
\hline
3.14 (0.98) &  0.9 (0.76) &   0 (0.0) &       1 &  1000 &           29.95 \\
\hline
\end{tabular}
\end{table}

\subsection{How is MCCE's performance when the prediction model is non-gradient based?} \label{non-gradient}

In addition to the restriction to categorical features with two levels, our method comparison required a gradient-based predictive model to be applicable to all the alternative methods. 
Again, MCCE is not restricted to gradient-based predictive models. 
Thus, below, we use MCCE to explain a random forest model on the Adult data set, to showcase that it is directly applicable to non gradient-based predictive models. 
We use a random forest model with 200 trees. 


The counterfactual method \texttt{C-CHVAE} is the only other on-manifold method in our list of benchmark methods that supposedly handles non-gradient-based models. Thus, it would have been interesting to compare the performance with this method. 
Unfortunately, a non-gradient-based version of this method is not currently implemented in \texttt{CARLA}.

The average performance metrics for MCCE are reported in Table \ref{tab:results_tree}. As in Section \ref{non-binarised}, the results are not directly comparable. Note, however, that the cost is quite similar to those for the ANN model. The computation time, on the other hand, is quite a bit higher for the random forest model. This is a result of increased prediction time (used when computing validity) for the random forest implementation compared to the ANN model.

\begin{table}[ht]
\caption{Average and standard deviation (in parentheses) of performance metrics for counterfactuals generated by MCCE when the predictive model is a \textbf{random forest}.
}
\label{tab:results_tree}
\centering
\begin{tabular}{ccccccc}
\hline
\multicolumn{7}{c}{Data set: Adult, $n_\text{test} = 1000$, $K = 1000$} \\
\hline
\hline
$L_0\downarrow$ & $L_1\downarrow$ &  violation$\downarrow$ & success$\uparrow$ & $N_{\text{CE}}\uparrow$ & time(s)$\downarrow$ \\
\hline
 2.48 (0.83) &  0.42 (0.43) &   0 (0.0) &       1 &  1000 &           36.53 \\
\hline
\end{tabular}
\end{table}

\subsection{Parameter values for competing methods}\label{parValues}
\begin{figure}[ht]
    \centering
    \includegraphics[width=1\linewidth]{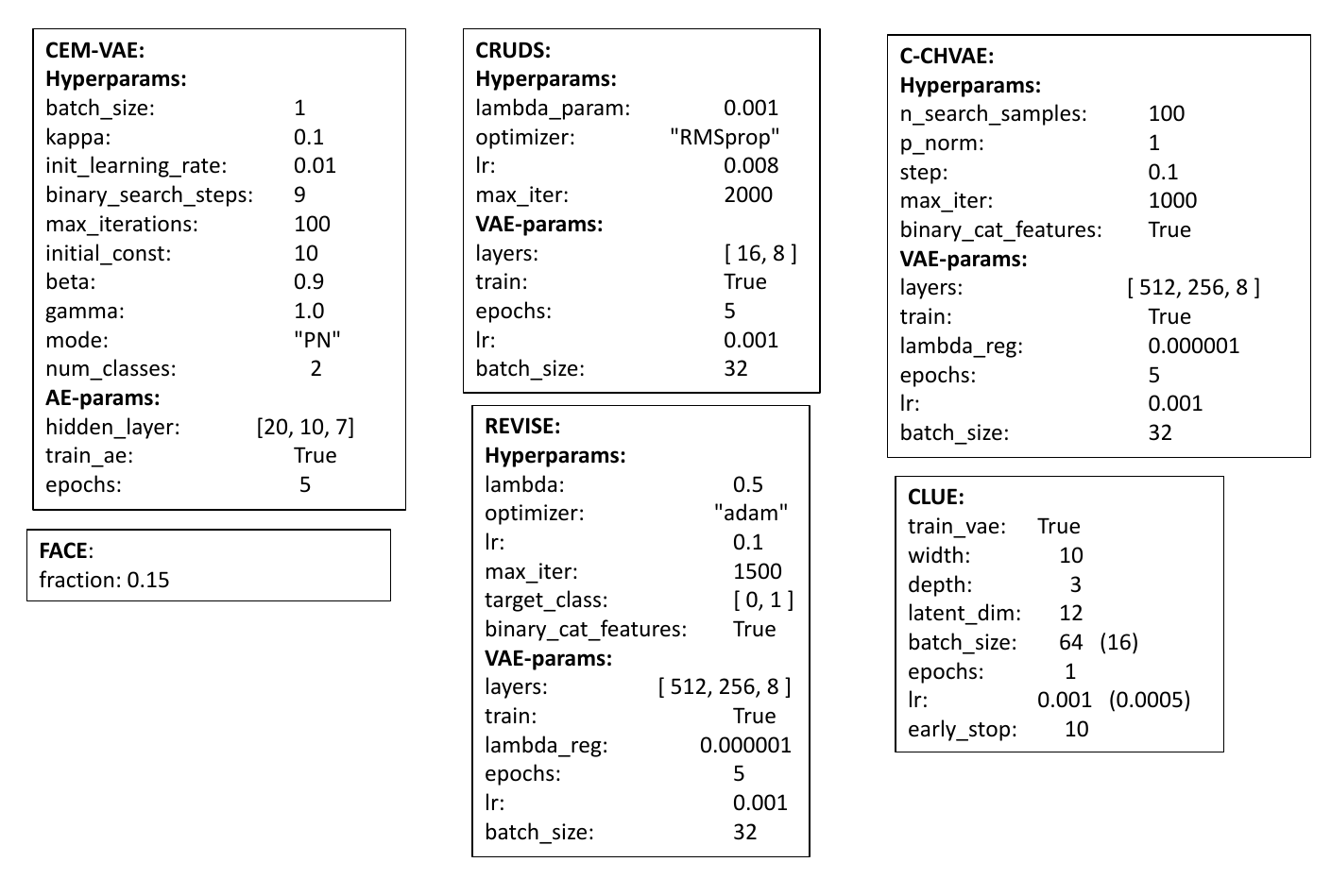} 
    \caption{Parameters used for the competing methods.}
    \label{fig:params}
\end{figure}

For all four data sets we used the default parameter values in \texttt{CARLA} for all the competing methods, except for \texttt{CLUE} where we had to use other values for the FICO and German Credit Data sets. The parameter values used for the different methods are shown in Figure \ref{fig:params}. The values in parenthesis are the  
ones used for the FICO and German Credit Data sets.

\subsection{How do the metrics and computation times compare when we \textit{do not} condition on the desired decision?}\label{sec:condition_response}

\begin{sidewaystable}
    \caption{Average and standard deviation (in parentheses) of performance metrics for counterfactuals generated with MCCE when \textbf{we do not condition on the decision}. 
    }
    \label{tab:timing_no_condition_response}
    \centering
    \begin{tabular}{rcccrcrrrr}
    \hline
    \multicolumn{9}{c}{Data set: Adult, $n_\text{test} = 1000$, $K$ varies} \\
    \hline
    \hline
    $K$ & $L_0\downarrow$ & $L_1\downarrow$ &  violation$\downarrow$ & $N_{\text{CE}}\uparrow$  & model(s)$\downarrow$  & gener(s)$\downarrow$ & post-proc(s)$\downarrow$ & total(s)$\downarrow$ \\
    \hline
    10 &   4.66 (1.4) &  2.19 (1.22) &   0 (0.0) &   633 &           3.87 &                1.09 &                   2.44 &            7.41 \\
    50 &  4.04 (1.27) &  1.71 (1.09) &   0 (0.0) &   837 &           3.87 &                1.59 &                   3.91 &            9.37 \\
   100 &  3.79 (1.24) &  1.50 (1.05) &   0 (0.0) &   875 &           3.87 &                1.77 &                   4.12 &            9.76 \\
  1000 &  3.19 (1.13) &  1.00 (0.85) &   0 (0.0) &   974 &           3.87 &                5.56 &                   4.92 &           14.35 \\
  5000 &   2.78 (0.9) &  0.69 (0.59) &   0 (0.0) &  1000 &           3.87 &               22.69 &                   6.20 &           32.76 \\
 \numprint{10000} &  2.64 (0.82) &  0.62 (0.54) &   0 (0.0) &  1000 &           3.87 &               43.95 &                   7.45 &           55.28 \\
 \numprint{25000} &  2.45 (0.81) &  0.55 (0.50) &   0 (0.0) &  1000 &           3.87 &              108.74 &                  10.95 &          123.56  \\

 \hline
    \end{tabular}
\end{sidewaystable}

One of the novel contributions of this paper is the modeling of the decision alongside the mutable features, to then condition on the desired decision (and the immutable features) when generating the data set of potential counterfactuals. 
The idea behind this notion is to bias the method toward generating samples that are more likely to yield the desired decision.
In this section, we investigate whether including the decision on the modeling and then conditioning on the desired decision really ensures a higher proportion of valid samples.
For the test observations in the Adult data set we generate counterfactuals \textit{without} conditioning on the decision for various $K$ and show the usual metrics and run times in Table \ref{tab:timing_no_condition_response}, to be compared to Table \ref{tab:timing_increase_K}.
As suspected, this variation of MCCE is much less effective in generating \textit{valid} counterfactuals ($N_\text{CE}$ in Table \ref{tab:timing_no_condition_response} is much smaller than in Table \ref{tab:timing_increase_K}). 

Furthermore, the fitting and generation times exhibit a slight decrease compared to the regular MCCE approach. Although one might have expected the training time to decrease due to fewer candidate features per split, the omission of the decision variable can result in larger trees as the available features convey less information. This was precisely the case for the given dataset, also leading to a slight increase in generation time.

On the other hand, the post-processing time is significantly reduced with this alternative approach, as it only generates around 25\% unique and valid samples (compared to 86\% with the normal MCCE), requiring fewer $L_0$ and $L_1$ calculations. In total, the computation time is smaller for the MCCE approach which does \textit{not} condition on the desired decision.

However, it is essential to highlight that this alternative approach requires a significantly higher value of $K$ in order to generate valid counterfactuals for all $1000$ test observations. When not conditioning on the desired decision, a $K$ value of $5,000$ is needed compared to only $50$ when conditioning on the desired response. Consequently, this negates the initial speed-up of the former method, making the original MCCE a clearly superior alternative.

\subsection{Quality of generated data}\label{sec:quality2}
In Section \ref{sec:quality} we showed the histograms for the generated data for four of the variables in the FICO data set. Here, we show the histograms for the rest of the variables. The histograms for the generated data are in white while the histograms for the real data are in dark grey.
As we can see, the marginal distributions of the generated data match the original ones very well also for these features.

\begin{figure}[ht]
    \centering
    \includegraphics[width=1\linewidth]{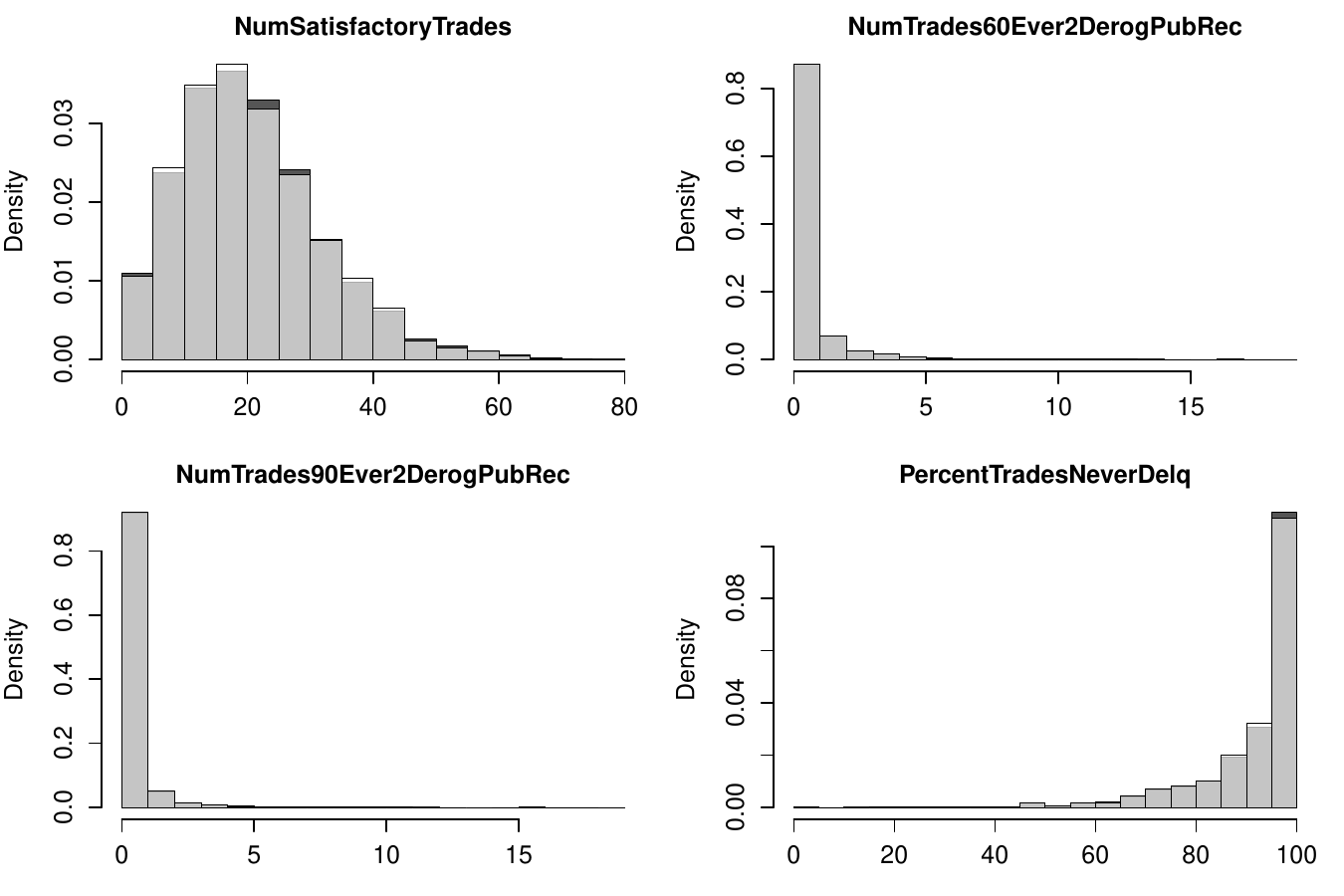} 
    \caption{Histograms for four of the variables in the generated data set (white) with the histograms for the real data superimposed (dark grey).
     Where the histograms overlap, the blend of white and dark grey gives a light grey color.}
    \label{fig:histograms1}
\end{figure}

\begin{figure}[ht]
    \centering
    \includegraphics[width=1\linewidth]{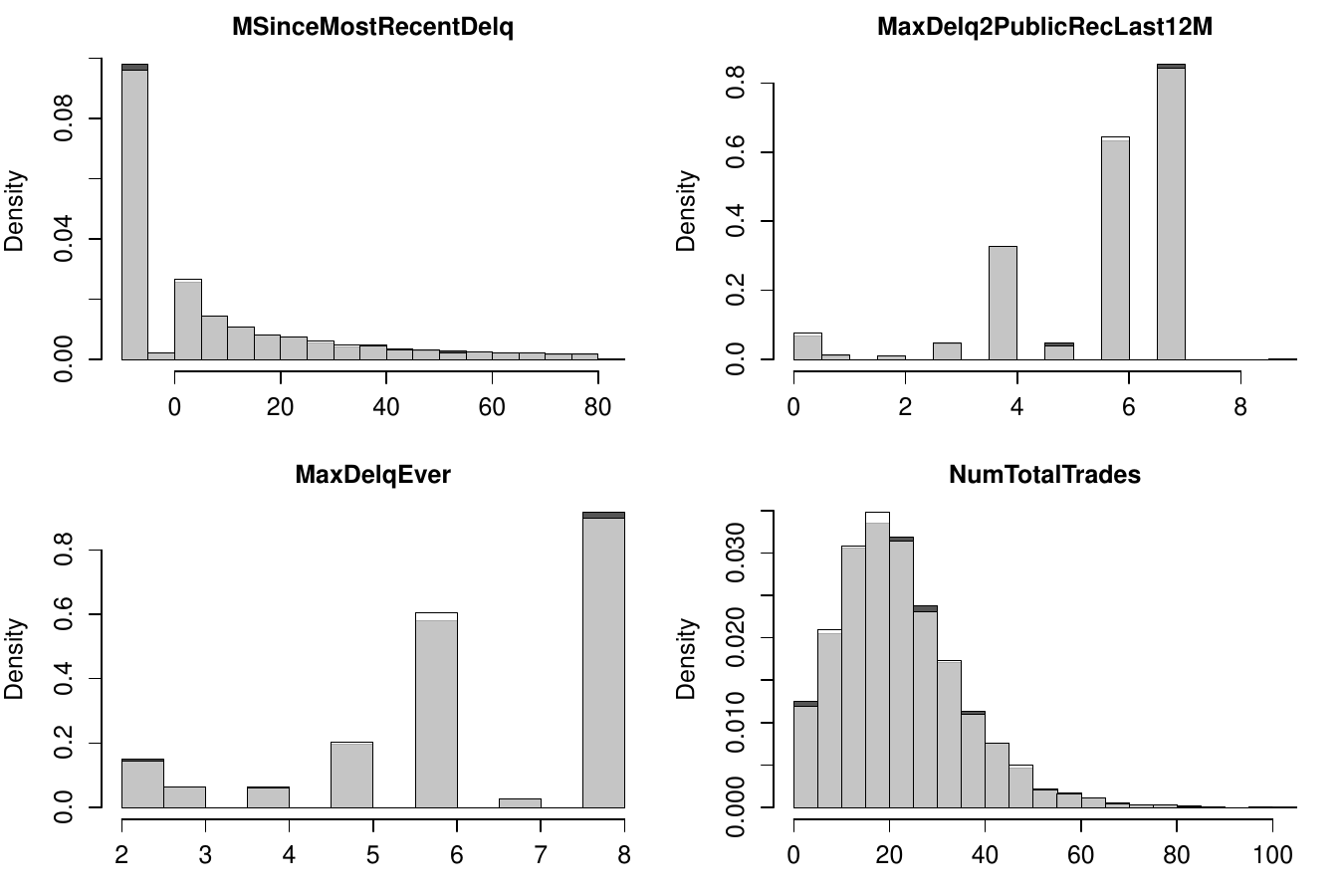} 
    \caption{Histograms for four of the variables in the generated data set (white) with the histograms for the real data superimposed (dark grey).
     Where the histograms overlap, the blend of white and dark grey gives a light grey color.}
    \label{fig:histograms2}
\end{figure}

\begin{figure}[ht]
    \centering
    \includegraphics[width=1\linewidth]{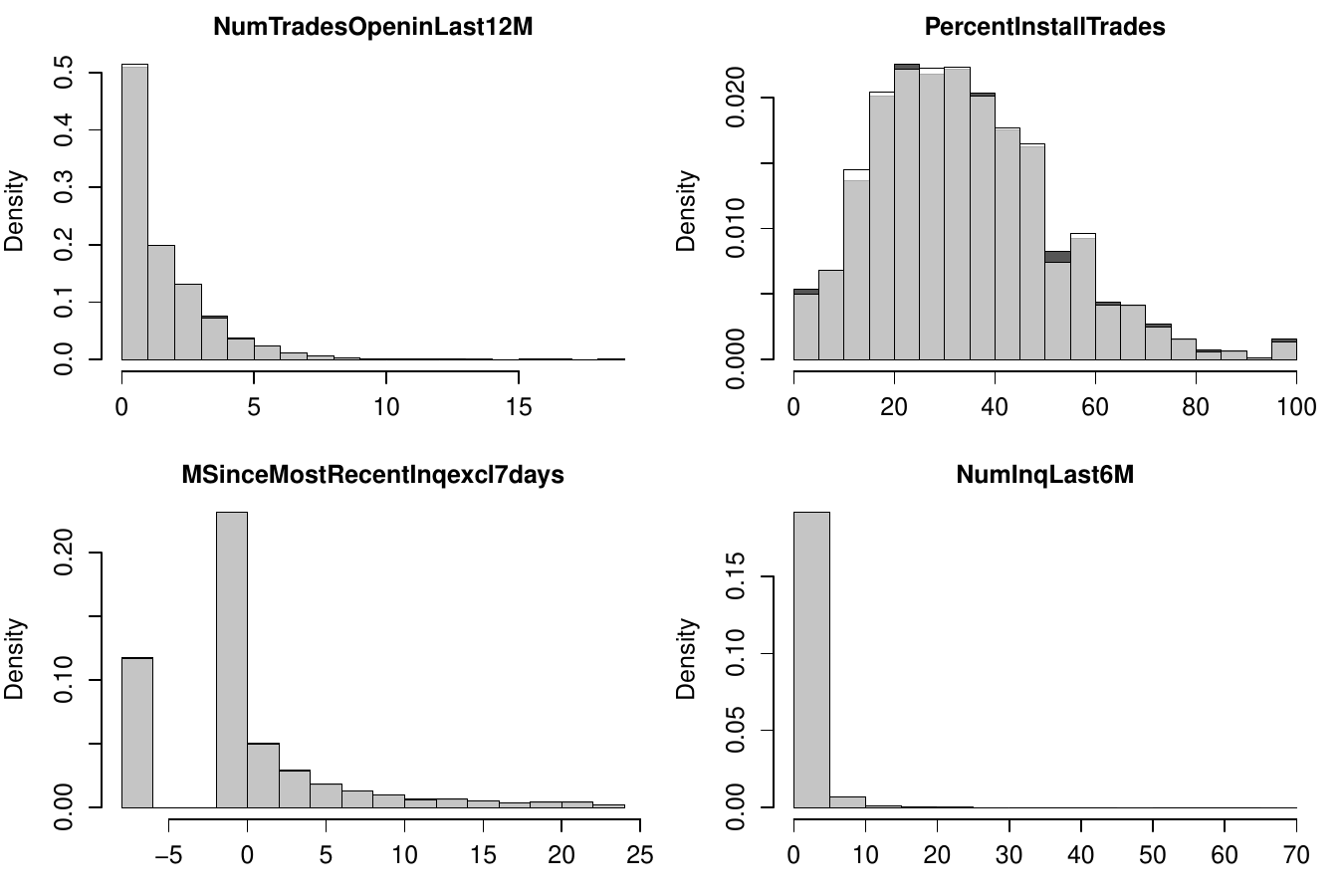} 
    \caption{Histograms for four of the variables in the generated data set (white) with the histograms for the real data superimposed (dark grey).
     Where the histograms overlap, the blend of white and dark grey gives a light grey color.}
    \label{fig:histograms3}
\end{figure}

\begin{figure}[ht]
    \centering
    \includegraphics[width=1\linewidth]{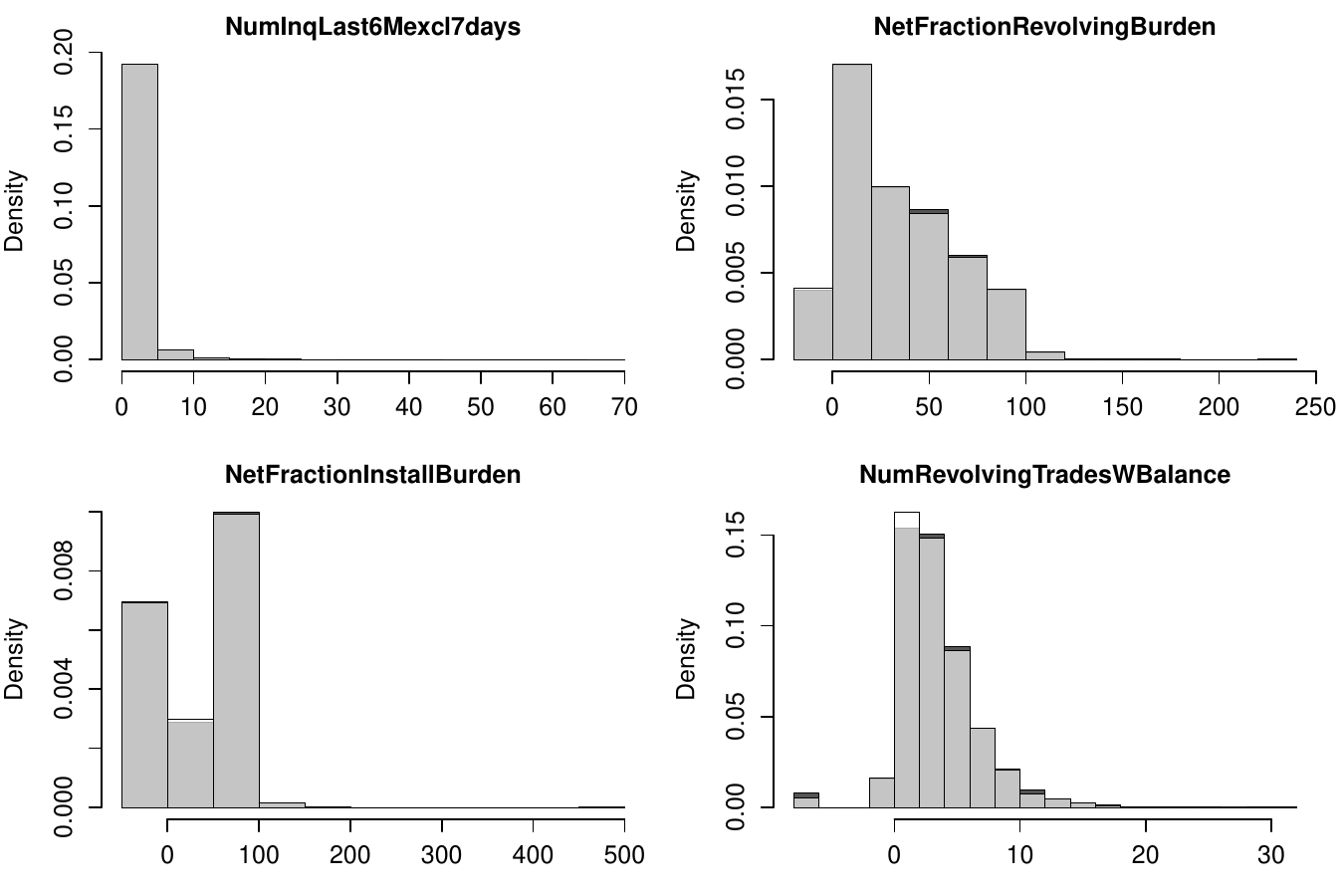} 
    \caption{Histograms for four of the variables in the generated data set (white) with the histograms for the real data superimposed (dark grey). Where the histograms overlap, the blend of white and dark grey gives a light grey color.}
    \label{fig:histograms4}
\end{figure}

\begin{figure}[ht]
    \centering
    \includegraphics[width=1\linewidth]{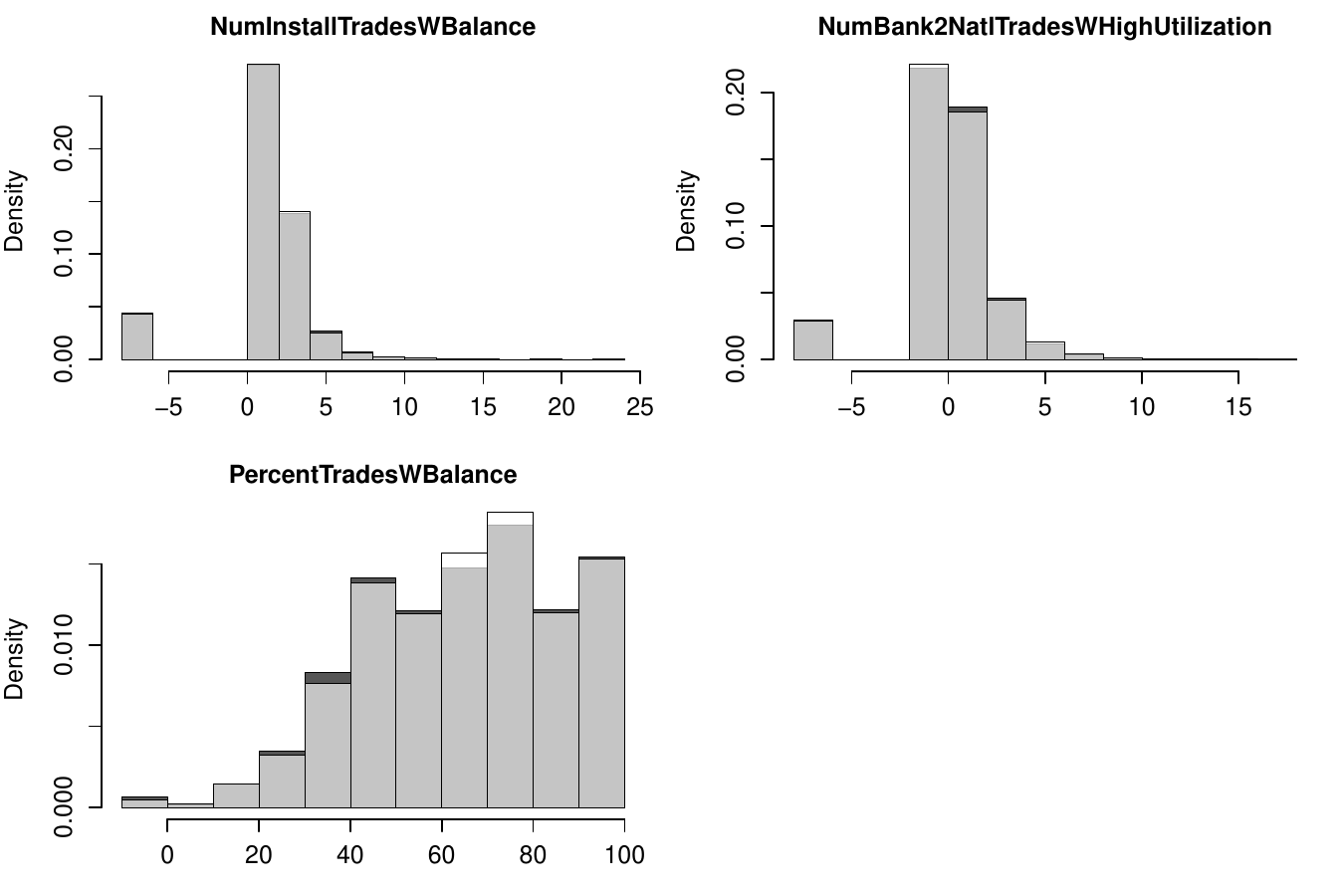} 
    \caption{Histograms for three of the variables in the generated data set (white) with the histograms for the real data superimposed (dark grey).
     Where the histograms overlap, the blend of of white and dark grey gives a light grey color.}
    \label{fig:histograms5}
\end{figure}

\subsection{Simulations mimicking real data experiments}\label{sec:real_data_sim}

In Section \ref{sec:limitations} we performed simulation experiments with a linear model when varying the dimension $p$ (with no fixed features, i.e.,~$u=0$), $n_{\text{train}}$, $n_{\text{test}}$ and $K$.
To showcase that these scalability results are relevant and valid in a broader context, we also ran simulations with quantities mimicking the four real data sets in Section \ref{subsec:experiment1}. As seen from the table, the total computation times in the simulation (27, 50, 7, and 49 seconds) are similar in magnitude to those recorded in the real data experiments (25, 32, 6, and 34 seconds), despite the model, the feature dependence and most likely the tree depth, differing significantly. This indicates that the scalability in our basic simulation study in Section \ref{sec:limitations} generalizes roughly to real data settings as well.

\begin{table}[ht!]
\caption{Computation times (in seconds) of the three steps of MCCE when mimicking $n_{\text{test}}$,  $n_{\text{train}}$, and $p/u/q$, with $K=1000$ for the four real data experiments in Section \ref{subsec:experiment1}. The displayed computation times represent the mean of 10 repeated computations.
}
\label{tab:datasimres}
\centering
\begin{tabular}{l|rrrr}
  \hline
mimicked dataset & model(s)$\downarrow$ & gener(s)$\downarrow$ & post-proc(s)$\downarrow$ & total(s)$\downarrow$  \\ 
  \hline
Adult &  10.28 & 9.60 & 7.30 & 27.18 \\ 
  GMC &  34.28 & 8.05 & 7.19 & 49.52 \\ 
  German credit & 0.56 & 5.38 & 1.46 & 7.40 \\ 
  FICO &  7.60 & 33.42 & 7.53 & 48.55 \\ 
   \hline
\end{tabular}
\end{table}


\clearpage





\end{appendices}


\bibliography{bibliography}


\end{document}